\documentclass[11pt]{article}

\usepackage[preprint]{acl}

\usepackage{times}
\usepackage{latexsym}

\usepackage[T1]{fontenc}

\usepackage[utf8]{inputenc}

\usepackage{microtype}

\usepackage{inconsolata}

\usepackage{graphicx}

\usepackage{subfiles}
\usepackage{xcolor}
\usepackage{makecell}

\usepackage{algorithm}
\usepackage{algpseudocode}
\usepackage{multirow}
\usepackage[normalem]{ulem}

\newcommand{\ps}{$\pm$}

\usepackage{tcolorbox} 
\newtcolorbox{fullwidthbox}{
    colback=gray!10, 
    colframe=black, 
    width=\textwidth, 
    sharp corners, 
    boxrule=1pt, 
    left=5pt, right=5pt, top=5pt, bottom=5pt 
}

\title{How Much Would a Clinician Edit This Draft?\\Evaluating LLM Alignment for Patient Message Response Drafting}

\author{Parker Seegmiller\textsuperscript{1}, Joseph Gatto\textsuperscript{1}, Sarah E. Greer\textsuperscript{1}, \\ \textbf{Ganza Belise Isingizwe\textsuperscript{1}, Rohan Ray\textsuperscript{1}, Timothy Burdick\textsuperscript{2,3}, Sarah M. Preum\textsuperscript{1}} \\
\textsuperscript{1} Department of Computer Science, Dartmouth College\\
\textsuperscript{2} Department of Community and Family Medicine, Dartmouth Health\\
\textsuperscript{3} The Dartmouth Institute, Dartmouth College \\
\texttt{\{pkseeg.gr, sarah.masud.preum\}@dartmouth.edu}}

\begin{document}
\maketitle

\begin{abstract}
Large language models (LLMs) show promise in drafting responses to patient portal messages, yet their integration into clinical workflows raises various concerns, including whether they would actually save clinicians time and effort in their portal workload. We investigate LLM alignment with individual clinicians through a comprehensive evaluation of the patient message response drafting task. We develop a novel taxonomy of thematic elements in clinician responses and propose a novel evaluation framework for assessing clinician editing load of LLM-drafted responses at both content and theme levels. We release an expert-annotated dataset and conduct large-scale evaluations of local and commercial LLMs using various adaptation techniques including thematic prompting, retrieval-augmented generation, supervised fine-tuning, and direct preference optimization. Our results reveal substantial epistemic uncertainty in aligning LLM drafts with clinician responses. While LLMs demonstrate capability in drafting certain thematic elements, they struggle with clinician-aligned generation in other themes, particularly question asking to elicit further information from patients. Theme-driven adaptation strategies yield improvements across most themes. Our findings underscore the necessity of adapting LLMs to individual clinician preferences to enable reliable and responsible use in patient-clinician communication workflows.
\end{abstract}


\section{Introduction}
\label{sec:intro}

\begin{figure}
    \centering
    \includegraphics[width=1.0\columnwidth]{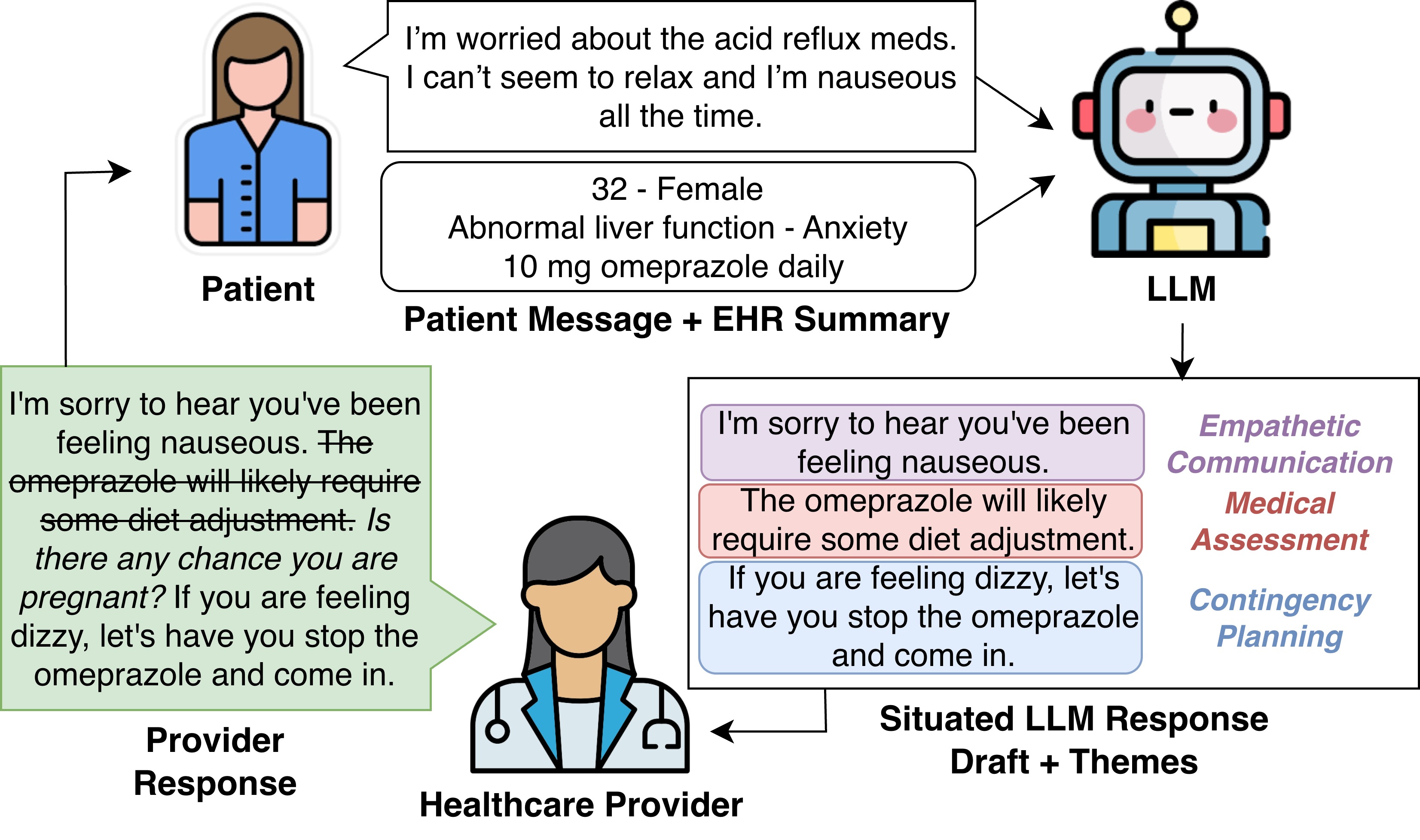}
      \caption{Patient message response drafting. LLMs draft responses to patient messages, then clinicians edit the draft by \sout{deleting} and \textit{adding} content as needed. We evaluate content-level and theme-level alignment between clinicians and LLMs.}
      \vspace{-10pt}
    \label{fig:task_overview}
\end{figure}



The use of large language models (LLMs) for \textbf{drafting responses to asynchronous patient messages} has garnered significant interest in the medical community \cite{hu2025systematic}. 
This would involve the integration of LLMs in the patient-clinician communication loop by drafting an initial clinician response to an incoming patient message, which the clinician would then edit and send to the patient. Figure \ref{fig:task_overview} shows an example of this task: generating a response draft to a patient-initiated message, given the message and a summary of the patient's relevant electronic health record (EHR) data. 
Responding to patient portal messages places a heavy burden on clinicians due to increasing use of the patient portal and significant clinical workforce constraints  \cite{Budd2023BurnoutRT, Underdahl2024PhysicianBE, Martinez2023PatientPM, yan2021exploring}. As such, there is growing interest in developing AI-mediated support for improving efficiency and engagement in patient portal messaging \cite{gatto2025fuqg, Biro2025OpportunitiesAR}. Thus, patient portal messaging is a high-stakes, real-world setting for evaluating LLMs on the task of drafting responses.

Prior work has gathered clinician feedback on LLM response drafts to patient portal messages with mixed results. Some studies report that these responses can be useful \cite{hu2025systematic, Garcia2024ArtificialID, bootsma2025ai, English2024UtilityOA}. However, there is evidence that LLM responses often diverge from clinician responses in style and content, and lack accuracy \cite{hu2025systematic, Biro2025OpportunitiesAR, Sharma2025EditingWA}. 

\begin{table*}
  \centering
  \small
  \begin{tabular}{lp{5.5cm}p{6.5cm}}\hline
    \textbf{Theme} &\textbf{Example Frame} &\textbf{Example Response Element} \\\hline
    \makecell[l]{Empathy} &Encouragement of patient treatment effort &You've been doing a great job with your tapering. \\\hline
    \makecell[l]{Symptom Questions} &Asking about location of symptoms &Has your pain only been in your lower back? \\\hline
    \makecell[l]{Medication Questions} &Asking about intake of medications &Have you been taking your Amoxicillan regularly? \\\hline
    \makecell[l]{Medical Assessment} &Explanation of test result &Your iron levels look normal. \\\hline
    \makecell[l]{Medical Planning} &Confirmation of required testing &Let's get you in for a bloodwork test. \\\hline
    \makecell[l]{Logistics} & Confirmation of clinic policy &We can only offer telehealth in the state. \\\hline
    Care Coordination &Promise of future patient contact &We'll reach out after we receive the results. \\\hline
    \makecell[l]{Contingency Planning} &Symptom-related backup plan &If you're feeling dizzy, please call triage \\
    \hline
    \end{tabular}
  \caption{Themes derived from clinician responses to patient portal messages, alongside representative frames and example response elements/utterances. For example, ``explanation of test result'' is a frame within the medical assessment theme, and ``your iron levels look normal'' is a clinician response component that falls under this frame. In total, we derive 8 clinician response themes comprised of 67 unique frames (examples in supplemental materials).}
  \vspace{-12pt}
  \label{tab:themes}
\end{table*}

Divergence between LLM response drafts and clinician responses may lead to either \textbf{unreliability}, if LLM response drafts must be significantly edited, contributing to clinicians' workload in responding to patient messages, or \textbf{irresponsibility}, if unedited low-quality LLM response draft elements are sent to the patient. Reliability is important, as clinicians spending significant time editing/improving the drafted response defeats the purpose of using LLMs to improve efficiency \cite{TaiSeale2024AIGeneratedDR, bootsma2025ai}. Clinician responsibility is critical, as LLM-generated drafts may contain clinically-significant errors and adversely impact the standards of care \cite{Biro2025OpportunitiesAR, Sharma2025EditingWA, chen2025retrieval}.



We investigate the use of LLM drafts in supporting clinician responses to patient messages, by evaluating alignment of LLMs to responses generated by real clinicians. Specifically, we aim to explore the content-level and theme-level alignment between clinician-written and LLM-generated responses, to inform responsible use of NLP in patient message response drafting. We answer three relevant research questions. 
\textbf{RQ1}: What constitutes a high-quality clinician response to a patient message?
\textbf{RQ2}: How might we automate evaluation of LLM response draft quality, with respect to clinician editing workload? 
\textbf{RQ3}: How can we adapt LLMs to support clinicians in generating quality responses to patient messages?

In answering these research questions, we make \textbf{four key contributions}. \textbf{First}, we use a clinicians-in-the-loop, hybrid approach to develop a clinically-relevant set of ``themes'' and frames to systematically characterize clinician responses to patient messages.
\textbf{Second}, we develop and validate a novel two-level evaluation framework for assessing clinician editing load given LLM-drafted responses to patient messages. 
\textbf{Third}, we annotate and release an expert-clinician-annotated dataset for evaluating performance on the patient message response drafting task\footnote{\url{https://hf.co/collections/PortalPal-AI/evaluating-alignment-for-patient-message-response-drafting}}. \textbf{Finally}, we conduct a rigorous evaluation of three local and three commercial LLMs on this task, using five LLM adaptation techniques varying in degree of supervision, finding that theme-driven adaptation of LLMs improves response drafting performance by 33\% over 0-shot models.

\section{Overview of Data}
\label{sec:data}

\label{sec:data_overview}
The patient-clinician conversations used in our experiments are collected from a large academic hospital in the United States. These conversations are sourced from the hospital's electronic health record (EHR) portal messaging platform. Patient portal messaging is an asynchronous healthcare communication service in which patients and their clinicians discuss a wide variety of patient health issues, including symptoms, medication efficacy, treatment planning, scheduling logistics, and more \cite{North2019ARA}. 

\begin{table*}
\small
  \centering
  \begin{tabular}{|c|cccccc|}\hline
    \textbf{Dataset} &\textbf{Source} &\textbf{Response} &\textbf{Clinician ct.} &\textbf{Size} &\textbf{Message} &\textbf{Response} \\\hline
    \textbf{IPPM} &Patient Portal Message + EHR &Theme-Guided &4 &300 &83\ps54 &53\ps32 \\
    \textbf{SyPPM} &Synthetic Message + EHR  &Theme-Guided &3 &100 &110\ps51 &70\ps26 \\
    \textbf{SoCPPM} &Patient Portal Message + EHR &Real-Time &196 &300 &69\ps45 &55\ps78 \\
    \hline
  \end{tabular}
  \caption{Summary of the three datasets. Patient messages in IPPM and SoCPPM, and EHR summaries for all datasets are sourced from a real EHR portal. SyPPM messages are semi-synthetic, generated using de-identified real patient messages for public release. Details on how clinician responses are collected and annotated are in Appendix \ref{app:redcap}. We include mean \ps standard deviation of the word count of patient messages and clinician responses.}
   \vspace{-12pt}
  \label{tab:dataset_summary}
\end{table*}


We begin with 610k total messages taken from the secure patient portal between 1/2020 - 9/2024. Our dataset includes messages from primary care, and thus includes a wide range of medical topics. We gather all patient-initiated messages which received a written clinician response to create 146k conversations, i.e. original patient message and response from a clinician. Our final data pool contains 10,105 unique patients, of which 64\% are female and 36\% are male, with ages ranging between 18-80. Each sample in our data pool consists of a patient message, a clinician response, and a summary of the patient's chart or electronic health record (EHR) data\footnote{See appendix \ref{app:dataset} for full dataset details}. We utilize 144k conversations from the data pool as training data, and gather evaluation datasets from the remaining 2k conversations.

\subsection{Thematic Analysis of Responses}
\label{sec:data_themes}

We address RQ1 by carefully deriving elements of high-quality clinician responses to patient messages. Based on manual thematic analysis of our real patient-clinician conversations, and research workshops with a team of 13 expert primary care physicians, nurses, and triage nurses, we derive a set of clinically-relevant ``themes'' which can be used to characterize the quality of clinician responses to patient messages \cite{braun2006using, sun2013messaging}. These themes can be found in Table \ref{tab:themes}. Appendix \ref{app:thematic_analysis} gives full details of our mixed-methods approach to identify these themes.

\subsection{Summary of Evaluation Datasets}

Table \ref{tab:dataset_summary} summarizes our three evaluation datasets. Here we briefly describe the three datasets derived from these 2k conversations and share additional dataset details in Appendix \ref{app:dataset_eval_datasets}.
Each sample in each dataset is a tuple of strings $\{m, c, r\}$ consisting of a patient message $m$, a summary $c$ of the patient's EHR chart and a single clinician response $r$. The Ideal Patient Portal Messaging (\textbf{IPPM}) dataset is created to evaluate LLMs in a setting where clinicians do not face the same resource constraints as in the real-world, thus responses are written by a team of paid expert clinicians who are guided by the themes derived in Section \ref{sec:data_themes}. The publicly-available Synthetic Patient Portal Message (\textbf{SyPPM}) contain semi-synthetic patient portal messages, paired with real de-identified patient EHR summaries, with responses collected via the same method as IPPM. The Standards of Care Patient Portal Messaging (\textbf{SoCPPM}) dataset is created to evaluate LLMs in a practical setting, where response drafts are compared with a clinician response which was sent via the portal in real time, thus responses are collected via the patient portal.


\section{Scalable Evaluation of LLMs}
\label{sec:evaluation}

\begin{figure*}
    \centering
    \vspace{-10pt}
    \includegraphics[width=0.8\textwidth]{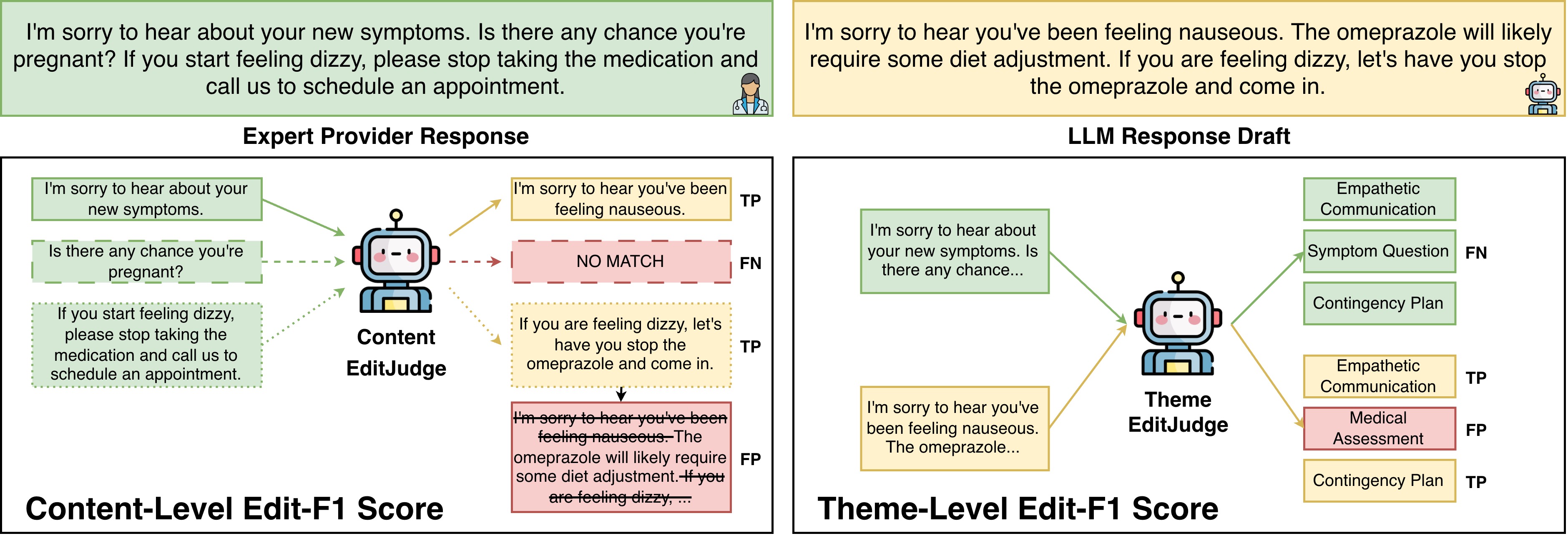}
    \caption{The EditJudge Evaluation Framework for evaluating LLM response drafts. The content-level edit-F1 score identifies matching content in the response draft ($EM$, i.e. true positives), along with expected deletions ($ED$, false positives) and expected additions ($EA$, false negatives) needed in order to align the LLM response draft with the clinician's desired response. The theme-level edit score identifies matching themes, serving as a relaxed evaluation of the theme-level alignment. }
    \vspace{-10pt}
    \label{fig:eval_framework}
\end{figure*}

We want to evaluate the reliability of LLM responses on the response drafting task (RQ2). Our evaluation seeks to identify: in order to achieve the same quality of response, 1) how much content would the clinician need to \textit{add} to the LLM draft? and 2) how much content would the clinician need to \textit{remove} from the LLM draft? Hence, we use a reference-based approach which directly compares an LLM draft with a response written by an expert clinician \cite{li2024llms}. Comparing what needs to be removed from and added to an LLM-drafted response to achieve an expert-written response, is analogous to measuring 1) \textit{recall}, i.e. how much of the expert-written response is covered by the LLM-drafted response, and 2) \textit{precision}, i.e. how much of the LLM-drafted response is matched in the expert's response. As our goal is to identify the editing load of a clinician using a LLM-as-judge framework, we call this the \textbf{EditJudge Evaluation Framework} (Figure \ref{fig:eval_framework}). This framework is a human-AI collaborative, task-specific, reference-based, LLM-as-judge evaluation framework \cite{li2025generation, bavaresco-etal-2025-llms}.

We use two measures of editing load to capture complementary aspects of alignment between generated and reference responses. The \textit{content-level edit-F1} score assesses whether a response drafting LLM reproduces specific clinical facts, instructions, or action items present in the reference, which is critical for safety and correctness. However, clinically appropriate drafts may vary substantially in wording or level of detail while addressing the same underlying intent. The \textit{theme-level edit-F1} score captures higher-level alignment by measuring whether the response addresses thematically similar clinical goals, concerns, and communicative functions (e.g., reassurance, triage guidance, or follow-up planning), even when the granular content differs. Using both metrics distinguishes incomplete response drafts from those that are semantically (content-level) and thematically aligned but phrased differently, providing a more reliable evaluation of response draft quality.  

\subsection{Content-Level edit-F1 Score}
\label{sec:exp_setup:content}

Given an expert-written clinician response $r_e$ and an LLM response draft $r_d$, the content-level edit-F1 score aims to identify how many \textit{expected additions} ($EA$) and \textit{expected deletions} ($ED$) are needed from the clinician, in order to unify $r_d$ with $r_e$. Matching content in the response draft $r_d$ is referred to as an \textit{expected match} ($EM$), meaning we would not expect the clinician to have to rewrite that content in order to achieve their desired response $r_e$, saving the clinician time and achieving reliability via LLM response drafting. 

We give our algorithm for counting $EA$, $ED$, and $EM$ in Algorithm \ref{alg:content_alg} in Appendix \ref{app:judge_model_details}. This algorithm splits an expert-written response $r_e$ into atomic elements (sentences), then for each element uses a fine-tuned judge LLM (content-level editJudge) to either identify expected matches $EM$ in the response draft $r_d$, or expected additions $EA$ to the response draft to achieve that element. The content-level editJudge takes as input a sentence from the expert-written response $s_e$ and the LLM-drafted response $r_d$, and outputs either the matching content from the LLM-drafted response $s_d$, or ``NO MATCH'' if there is no matching content. Finally, this algorithm identifies expected deletions $ED$ in the response draft by quantifying the remaining amount of unmatched content. By treating expected matches, expected additions, and expected deletions as true positives, false negatives, and false positives respectively, we calculate recall, i.e. the percentage of the expert-written response $r_e$ which does not need to be added to $r_d$, and precision, i.e. the percentage of the LLM response draft $r_d$ which does not need to be removed. We calculate the harmonic mean of the content-level recall and precision scores (i.e. $F_1$) and call this the \textbf{content-level edit-F1 score}. Assuming additions and deletions are evenly-weighted, content-level edit-F1 gives the expected reduction in editing load for the clinician by using the LLM response draft.

We evaluate 10 variations of content-level editJudge models, and select a fine-tuned LLama-3-8B-Instruct model for use in our experiments in Section \ref{sec:exp_results}. This editJudge model achieves 96\% agreement with expert human annotators, including 92\% overlap with expert-annotated matching content decisions. We discuss data annotation, training, and evaluation of editJudge models in Appendix \ref{app:judge_model_details}.

\subsection{Theme-Level edit-F1 Score}
\label{sec:exp_setup:themes}

Given a clinician response $r_e$ and an LLM response draft $r_d$, the theme-level edit-F1 score aims to identify the higher-level themes in the clinician response $r_e$ which are correctly matched by the themes in the LLM response draft $r_d$. To identify themes in each response, we develop and evaluate a theme-level editJudge model. Given a sentence from either the clinician response $s_e \in r_e$ or the LLM drafted response $s_d \in r_d$, the theme-level editJudge model assigns a theme label $l_s$. Predicting clinician response themes is an 9-class multi-label classification task, as there are 8 high-level themes (see Table \ref{tab:themes}) and an ``Other'' class to capture miscellaneous themes not captured in the main 8 classes. Using the theme labels $l_{s_d}$ assigned to sentences $s_d$ from the LLM drafted response $r_d$ as predictions for the theme labels $l_{s_e}$ assigned to sentences $s_e$ from the clinician response $r_e$, the \textbf{theme-level edit-F1 score} is the micro average $F_1$ of theme predictions. We develop and evaluate a fine-tuned theme-level editJudge theme classification model which achieves an $F_1$ score of 0.82 on expert-annotated dataset (details in Appendix \ref{app:judge_model_details}).



\section{Experimental Setup}
\label{sec:exp_setup}

We are interested in how LLMs might be more closely aligned with expert clinicians, to increase the reliability and responsibility of LLMs in response drafting (RQ3). We describe the models used in our evaluation, a measure of inter-annotator predictability (IAP) to contextualize our results, and a measurement of theme frequency in clinician and LLM response drafts.

\subsection{Models and Adaptation Methods}
\label{sec:eval_models}

\subsubsection{Local and Frontier LLMs}
\label{sec:eval_models_local}

Locally-hosted LLMs are often preferable in clinical settings due to the sensitive nature of protected health information (PHI) and the frequency with which PHI occurs in patient portal messages \cite{sallam2023chatgpt, zhou2023skingpt}. Token throughput and hosting memory constraints are also important considerations \cite{lorencin2025optimizing}. As such, we are interested in evaluating 7-8b parameter LLMs on the response drafting task. We use three models: (i) the instruction-tuned Llama3-8B model \cite{llama3modelcard}, (ii) a healthcare-specific version of the same model Aloe-8B \cite{gururajan2024aloe}, and (iii) Qwen3-8B \cite{qwen3technicalreport} from a different model family. We also test three commercial models on the SyPPM dataset, our public dataset: (i) Claude 4.5 Sonnet \cite{claude45sonnet}, (ii) Gemini 2.5 Pro \cite{comanici2025gemini25pushingfrontier}, and (iii) GPT-OSS \cite{agarwal2025gpt}. 



\subsubsection{Adaptation Techniques}
\label{sec:exp_setup_adapts}
We are interested in exploring several avenues for aligning LLMs with expert clinicians to improve reliability and responsibility. We briefly describe each adaptation strategy here, providing full details in Appendix \ref{app:adaptation_details}, and prompts in Appendix \ref{app:prompts}.

\textbf{0-Shot}. Minimally-guided responses from each model are evaluated to identify how closely-aligned the LLM is with expert clinicians.

\textbf{Thematic}. Some prior work has shown that prompting techniques can improve LLM performance on patient messaging tasks \cite{Genovese2025ArtificialIF}. We are interested in whether the themes derived in Section \ref{sec:data_themes} can align LLMs more closely with expert clinicians. The thematic prompt includes a brief explanation of each of the 8 themes, to guide the LLM with context.

\textbf{RAG}. Retrieval augmented generation has been used in other patient messaging tasks to improve style and content of LLM responses \cite{chen2025retrieval}. We perform 5-shot RAG prompting. 

\textbf{SFT}. Supervised fine-tuning on prior patient-clinician conversations has proven to be an effective way to adapt LLM for patient message response drafting \cite{liu2024leveraging}. We perform SFT using all 144k training messages. 

\begin{table*}[h]
\small
\centering
    \begin{tabular}{|c|c|ccc|ccc|}
    \hline
    \multicolumn{2}{|c|}{} & \multicolumn{3}{c}{\textbf{Content-Level}} & \multicolumn{3}{|c|}{\textbf{Theme-Level}} \\ \hline
    Dataset & Model & Precision & Recall & Edit-F1 & Precision & Recall & Edit-F1 \\ \hline
    \multirow{5}{*}{\textbf{IPPM}}
    &0-Shot & 0.07\ps0.02 & 0.26\ps0.04 & 0.10\ps0.02 &0.49\ps0.03 &0.74\ps0.03 &0.58\ps0.02 \\ 
    &Theme & 0.06\ps0.01 & \textbf{0.30\ps0.05} & 0.09\ps0.01 &0.47\ps0.01 &\textbf{0.80\ps0.02} &0.58\ps0.01 \\ 
    &RAG & 0.11\ps0.03 & 0.30\ps0.17 & 0.13\ps0.01 &0.48\ps0.20 &0.66\ps0.09 &0.56\ps0.02 \\ 
    &SFT & \textbf{0.15\ps0.01} & 0.16\ps0.00 & \textbf{0.14\ps0.01} &\textbf{0.64\ps0.01} &0.57\ps0.01 &\textbf{0.60\ps0.01} \\ 
    &TADPOLE & 0.13\ps0.01 & 0.18\ps0.01 & \textbf{0.14\ps0.01} &0.54\ps0.00 &0.65\ps0.02 &0.59\ps0.01 \\ \hline
    \multirow{6}{*}{\textbf{SyPPM}}
    &0-Shot & 0.12\ps0.04 & 0.31\ps0.03 & 0.16\ps0.04 &0.47\ps0.02 &0.46\ps0.03 &0.47\ps0.02 \\ 
    &Theme & 0.11\ps0.00 & \textbf{0.33\ps0.10} & 0.15\ps0.02 &0.50\ps0.01 &\textbf{0.58\ps0.01} &0.54\ps0.00 \\ 
    &RAG & 0.17\ps0.08 & 0.28\ps0.07 & 0.18\ps0.05 &0.47\ps0.03 &0.43\ps0.02 &0.45\ps0.02 \\ 
    &SFT & \textbf{0.22\ps0.01} & 0.17\ps0.01 & 0.18\ps0.0 &\textbf{0.64\ps0.01} &0.41\ps0.01 &0.50\ps0.01 \\ 
    &TADPOLE & 0.21\ps0.01 & 0.20\ps0.02 & \textbf{0.20\ps0.01} &0.62\ps0.01 &0.54\ps0.02 &\textbf{0.58\ps0.01} \\ \cline{2-8}
    &\textit{Gemini} &\textit{0.20} &\textit{0.43} &\textit{0.26} &\textit{0.58} &\textit{0.69} &\textit{0.64} \\ \hline
    \multicolumn{2}{|c|}{\textit{IAP}} & \textit{0.26} & \textit{0.25} & \textit{0.24} &\textit{0.61} &\textit{0.63} &\textit{0.62} \\ \hline
    \end{tabular}
    \caption{Edit-F1 scores for LLM adaptations on the IPPM and SyPPM patient message response drafting datasets. Each model adaptation is performed on three underlying LLMs, we report scores as average\ps standard deviation. We report content-level precision, recall, and edit-F1 (Section \ref{sec:exp_setup:content}), as well as theme-level precision, recall, and edit-F1 (Section \ref{sec:exp_setup:themes}). We include the best commercial model (Gemini + theme prompting) scores on the publicly-available SyPPM dataset. Finally, we report content-level inter-annotator predictability (IAP), comparing LLM performance and expert human alignment.}
    \vspace{-12pt}
    \label{tab:combined_results}
\end{table*}


\textbf{TADPOLE}. We develop a novel Thematic Agentic Direct Preference Optimization for Learning Enhancement strategy for creating theme-driven preference training data for DPO \cite{rafailov2023direct}. TADPOLE uses response enhancement agents designed for each theme derived in Section \ref{sec:data_themes}. We test several preference pair creation strategies (details in Appendix \ref{app:adaptation_tadpole}), and report the results of models trained using the best-performing strategy. 



\subsection{Inter-Annotator Predictability}
\label{sec:eval_iaa}
A key consideration when evaluating the LLM-clinician alignment is how closely-aligned clinicians are with each other. Clinician alignment may vary based on experience factors (e.g. role, years of experience, specialty), personality factors (e.g. writing style), and interpersonal factors (e.g. relationship with the patient). We gather 3 expert responses to 40 samples from the SyPPM dataset to quantify inter-annotator predictability (IAP). We calculate IAP using the editJudge framework to compare inter-human alignment on patient message response drafting. IAP gives us a measure of how useful a different clinician's response might be when used as a response draft. We also report inter-annotator agreement of ground truth in Tables \ref{app:tab:iaa_strict}-\ref{app:tab:iaa_cosine} in Appendix \ref{app:iaa}.

\subsection{Estimated Theme Frequency}
\label{sec:exp_setup_theme_freq}

As manually annotating sentence themes in all responses would be inefficient, we use our empirically-validated sentence-level theme classifier (theme-level editJudge LLM, achieves 0.82 F1 on test set in Appendix \ref{app:judge_model_details}) to classify themes in all clinician responses (i.e., ground truth) and all LLM response drafts to estimate thematic tendencies (see Table \ref{tab:theme_freq}).

\section{Results}
\label{sec:exp_results}

We evaluate six LLMs and five adaptation techniques on the IPPM and SyPPM response drafting evaluation datasets and discuss our findings. Due to space constraints, we discuss results on the SoCPPM dataset in Appendix \ref{app:additional_results}. 

Table \ref{tab:combined_results} contains both content-level and theme-level edit-F1 scores, \textbf{averaged} across the three local LLMs described in Section \ref{sec:eval_models_local}, alongside standard deviation. Table \ref{tab:combined_frontier} contains content- and theme-level edit-F1 scores for Claude 4.5 Sonnet, Gemini 2.5 Pro, and GPT-OSS reasoning models, using both 0-shot and thematic prompting adaptation. In Tables \ref{tab:combined_results} and \ref{tab:combined_frontier} we report micro average precision, recall, and edit-F1 at the content and theme levels. Table \ref{tab:theme_freq} contains theme frequencies for clinician responses and adapted LLM drafts, averaged across all evaluation datasets.  

\subsection{Content-Level Results}
\label{sec:exp_results:content}

\textbf{Usefulness of Thematic Context:} We find that fine-tuned models achieve highest precision, theme-prompted models achieve highest recall, and the TADPOLE adaptation strategy offers the best blend of precision and recall with the highest average content-level edit-F1 scores. We find that added context improves LLM alignment with individual clinicians, and that edit-F1 performance generally scales with the amount of added context. Examining theme-specific content-level recall (Table \ref{app:tab:content_classwise} in Appendix \ref{app:additional_results_class_average}), TADPOLE-adapted models blend precision with empathetic communication content (0.30 average recall vs 0.28 average among other adaptations) and contingency planning content (0.27 vs 0.21)---two themes which tend to appear more in ``ideal'' response drafts. Among commercial models, thematic prompting adaptation improves performance of all three LLMs. We find that the best frontier-level model in our evaluation is Gemini 2.5 Pro adapted with thematic prompting, achieving 0.26 content-level and 0.64 theme-level edit-F1. Our single best-performing TADPOLE model (Qwen3-8B trained on the ``corrupted'' preference pairs\footnote{See TADPOLE results in Table \ref{app:tab:tadpole_content} in Appendix \ref{app:adaptation_tadpole}}) achieves comparable performance (0.25 content-level edit-F1 score) to the best-performing frontier model (Gemini 2.5 Pro + theme prompt, 0.26). Our evaluation suggests that using one of these models in patient message response drafting would lead to a 25-26\% reduction in clinician edits. 



\textbf{Epistemic Uncertainty:} Individual variation stemming from epistemic uncertainty is often observed in medicine  \cite{han2021physicians}, including patient message response drafting \cite{chen2024effect, Garcia2024ArtificialID, laukka2020health, baxter2024generative, english2024utility}. Our results support this finding (see Tables \ref{app:tab:iaa_strict}-\ref{app:tab:multiresponse} in Appendix \ref{app:iaa}). When one clinician's responses are used as drafts for another clinician, we find an average content-level edit-F1 score of 0.24---meaning that using another clinicians response as a draft only reduces clinician edits by 24\%. This indicates substantial epistemic uncertainty at the content level of clinician responses, i.e., LLMs specialized at the task level are subject to performance loss due to inter-clinician variation in judgment and preferences. This highlights the need for LLMs to be specialized at the expert level in order to further improve clinician efficiency with response drafts.

\begin{table*}
  \small
  \centering
  \begin{tabular}{|c|c|ccc|ccc|}\hline
    \textbf{} &\textbf{} &\multicolumn{3}{c}{\textbf{Content-Level}} &\multicolumn{3}{|c|}{\textbf{Theme-Level}} \\\hline
    \textbf{Prompt} &\textbf{Model} &\textbf{Pr} &\textbf{Re} &\textbf{Edit-F1} &\textbf{Pr} &\textbf{Re} &\textbf{Edit-F1} \\\hline
    \multirow{4}{*}{0-Shot} &GPT &0.03 &0.21 &0.05 &0.45 &\textbf{0.64} &0.53 \\
    &Gemini &0.17 &\textbf{0.40} &0.23 &0.52 &0.56 &\textbf{0.54} \\
    &Claude &\textbf{0.20} &0.38 &\textbf{0.25} &\textbf{0.52} &0.54 &0.53 \\
    &\textit{Avg} &\textit{0.13} &\textit{0.33} &\textit{0.18} &\textit{0.50} &\textit{0.58} &\textit{0.53} \\\hline
    \multirow{4}{*}{Theme} &GPT &0.06 &0.30 &0.09 &0.49 &\textbf{0.77} &0.60 \\
    &Gemini &\textbf{0.20} &\textbf{0.43} &\textbf{0.26} &0.56 &0.69 &\textbf{0.64} \\
    &Claude &0.16 &0.37 &0.22 &\textbf{0.58} &0.69 &0.63 \\
    &\textit{Avg} &\textit{0.14} &\textit{0.37} &\textit{0.19} &\textit{0.54} &\textit{0.72} &\textit{0.62} \\\hline
    \multicolumn{2}{|c|}{\textit{IAP}} & \textit{0.26} & \textit{0.25} & \textit{0.24} &\textit{0.61} &\textit{0.63} &\textit{0.62} \\
    \hline
    \end{tabular}
  \caption{Edit-F1 results for Claude 4.5 Sonnet, Gemini 2.5 Pro, and GPT-OSS reasoning models on the publicly-available SyPPM evaluation dataset. We evaluate each model using 0-shot and thematic prompts, and average scores for each prompt. We report precision, recall, and edit-F1 at both the content and theme levels. We report content-level IAP, comparing LLM performance and expert human alignment at the content level.}
  \label{tab:combined_frontier}
\end{table*}


\subsection{Theme-Level Results}
\label{sec:exp_results:themes}




\textbf{LLMs Generate Quality Empathetic Content}: Evaluating at the theme level shows that LLMs are capable of generating some themes accurately, while other themes are more challenging. For example, LLMs tend to generate the empathetic communication theme frequently (Table \ref{tab:theme_freq}), and they perform well overall at generating this theme---e.g. TADPOLE-adapted models achieve an average theme-level edit-F1 score of 0.99 on the empathetic communication theme in SyPPM (see Table \ref{app:tab:theme_classwise} in Appendix \ref{app:additional_results_class_average}). This finding supports \citet{English2024UtilityOA}, which finds that nurses report that LLM response drafts improve empathy and tone. On the contrast, Table \ref{tab:theme_freq} shows that unaligned LLMs will rarely ask follow-up questions. Unaligned LLMs tend to be misaligned with clinicians on question asking themes---e.g. 0-shot models achieve only 0.17 and 0.08 average theme-level edit-F1 scores on SyPPM symptom and medication question-asking themes (Table \ref{app:tab:theme_classwise}). Contextual adaptation greatly improves LLM performance at question asking, with TADPOLE-adapted LLMs improving to 0.79 and 0.49 average theme-level edit-F1 scores on SyPPM symptom and medication question-asking themes. 

\textbf{Individuality of Expert Clinicians:} In general, IAP is much higher at the theme level than at the content level, indicating that theme-level alignment is a more achievable goal when drafting clinician responses. However, some individual themes have very low IAP, e.g. treatment planning (0.07 IAP theme-level edit-F1 score in Table \ref{app:tab:theme_classwise}) and contingency planning (0.06). Discussions with various clinicians, including our annotators, highlight that different clinicians tend to think differently about how content will be perceived by patients -- e.g. some clinicians indicate that the benefits of providing contingency plans do not outweigh the burden it places on patients. This again underscores the need for LLMs to be able to be adapted at an individual level, in order to draft useful responses for individual clinicians with different roles (triage nurse, medical assistant, residents), specialties (internal medicine vs family medicine), years of experiences, and preferences. Individual alignment is vital for \textit{reliable} and \textit{responsible} use of LLM-mediated tools in high-stakes professional workflows like healthcare.


\subsection{Implications of Results}
\label{sec:implications}

\textbf{Reliable LLM Adaptation:} We find that unadapted LLMs tend to generate medical assessment themes more successfully than contextually-adapted LLMs. This is supported by our estimate of theme proportions (Table \ref{tab:theme_freq}), which finds that unadapted LLMs generate far more medical assessment and treatment planning themes than clinicians and contextually-adapted LLMs. These themes cover utterances related to medical decision making and communication, i.e., explaining test results, symptoms, and potential diagnoses; and recommending various forms of treatment. Intuitively, unadapted LLMs generate these themes more frequently as they relate to general LLM alignment principles, e.g., safety and helpfulness \cite{ji2023beavertails}. However, such behavior can lead to over-diagnosis and over-treatment \cite{kale2018overdiagnosis}, an emerging concern about using AI in medicine \cite{scott2024achieving}. Responses drafted by unadapted models also tend to be longer \cite{Garcia2024ArtificialID, hu2025systematic, TaiSeale2024AIGeneratedDR}, which may introduce more cognitive burden for clinicians, defeating the purpose of saving clinicians' time spent in responding to messages.


\begin{table*}
  \centering
  \begin{tabular}{|l|ccccccccc|}\hline
    \textbf{Response} &\textbf{Emp} &\textbf{Sym Q} &\textbf{Med Q} &\textbf{Assess} &\textbf{Plan} &\textbf{Logis} &\textbf{Coord} &\textbf{Cont} &\textbf{Oth} \\\hline
    Clinicians &0.85 &0.36 &0.30 &0.34 &0.19 &0.56 &0.45 &0.22 &0.02 \\\hline
    0-Shot &0.94 &0.02 &0.05 &0.89 &0.82 &0.59 &0.78 &0.18 &0.14 \\
    Theme &0.95 &\textbf{0.26} &0.13 &0.94 &0.79 &0.64 &0.82 &0.18 &0.20 \\
    RAG &\textbf{0.77} &0.01 &0.05 &0.79 &0.65 &\textbf{0.56} &\textbf{0.69} &0.11 &0.19 \\
    SFT &0.97 &0.02 &0.02 &0.23 &0.26 &0.38 &\textbf{0.69} &0.02 &\textbf{0.02} \\
    TADPOLE &0.99 &0.29 &\textbf{0.20} &\textbf{0.28} &\textbf{0.31} &0.36 &0.83 &\textbf{0.25} &0.01 \\
    \hline
  \end{tabular}
  \caption{Proportion of responses containing different thematic content, found in responses written by clinicians and various model adaptations. Clinician theme proportion is averaged across the IPPM, SyPPM, and SoCPPM datasets. LLM adaptation theme proportion is averaged over the three underlying LLMs as well as the three datasets. Bold proportions highlight the adaptation that was closest to clinician proportions.}
  \label{tab:theme_freq}
\end{table*}


\textbf{Importance of Evaluation:} Our evaluation measures how many edits a clinician would make to the LLM-generated draft before sending the response.
This is different from the goal of measuring response quality along pre-defined axes, and influences our decision to define a ground truth as a single clinician response, rather than a strategy such as rubric-based evaluation \cite{arora2025healthbench} or surveying expert feedback \cite{liu2024leveraging} on a generated response. Results from our targeted evaluation highlight the challenge of aligning models with individual clinicians' judgment, tone, and preferences when responding to patients. 
It also yields insights for future work to explore alternatives to response drafting to improve clinician efficiency, e.g., suggesting clinicians theme-based ``nudges'' --- rather than content--- for themes with higher epistemic uncertainty. 



\section{Related Works}
\vspace{-8pt}

\paragraph{Patient Message Response Drafting.}
Several works have studied the usefulness of LLMs in drafting clinician responses to patient messages. Most evaluate drafts via only clinician feedback, limiting the scale of evaluation, and employ only 0-shot frontier-level LLMs (most commonly OpenAI GPT-4) \cite{Biro2025OpportunitiesAR, Sharma2025EditingWA, English2024UtilityOA, small2024large, TaiSeale2024AIGeneratedDR, hu2025systematic, bootsma2025ai}. Our work extends prior work in two ways: (1) large scale evaluation of adapted LLMs and (2) inclusion of EHR data with message to situate generated responses. Results from prior studies are mixed, with some showing the potential of LLM drafts in promoting empathy and giving health advice \cite{English2024UtilityOA, eschler2015designing}, while others show that there is room for improvement in LLM draft completeness, tone, and simplicity \cite{Garcia2024ArtificialID, small2024large, chen2024effect}. Among studies that go beyond 0-shot evaluation, \citet{hu2025systematic} and \citet{kim2024perspectives} explore prompting strategies to improve LLM response drafts. Our thematic prompting strategy builds on prior work by incorporating a more granular-level understanding of LLM behavior in response generation across the constituent themes of a clinician's response. \citet{liu2024leveraging} is perhaps most similar to our work in that they perform SFT of a Llama model and evaluate on a small test set (n=10) using clinician feedback and BERTScore. Our work in developing a thorough automated evaluation framework aims to build on this by enabling larger-scale automated evaluation. Our focus on large-scale evaluation enables deeper insight into the risks and benefits of LLM use in patient message response drafting. 




\paragraph{Evaluation based on LLM-As-Judge.}


The use of LLMs as judges of LLM-generated content has grown significantly in recent years \cite{li2024llms, lin2023llm, li2025generation, bavaresco-etal-2025-llms}, including in healthcare text generation contexts \cite{croxford2025automating, Bedi2025MedHELMHE, Zhao2025AutomatingEO, krolik2024towards}. Perhaps most similar to our work, \citet{croxford2025automating} introduce an LLM-as-Judge framework for evaluating generated EHR summaries and use a rubric-based evaluation. In contrast, our novel edit-F1 framework is designed to estimate edit load, i.e., expected deletions/additions to LLM-generated draft. 

\section{Conclusion}
\label{sec:conclusion}

We have evaluated LLMs on the patient message response drafting task. We have developed a set of clinician response themes and used these to develop a novel evaluation framework for assessing clinician editing load given LLM response drafts. We have performed a large-scale evaluation of contextually-adapted LLMs and frontier LLMs, finding that contextual adaptation improves LLM performance. We highlight that individual clinician preferences vary significantly, and that adaptation of LLMs to individual clinicians is required to further increase the reliability and responsibility of LLM use for patient message response drafting.
\section{Limitations}
\label{sec:limitations}

\textbf{Dataset} Our data is drawn from a single hospital system and patient portal platform, which may limit generalizability to other healthcare settings with different workflows, patient populations, and communication norms. This is a rural hospital system. Future work may explore safety, bias and robustness of adapted LLMs in such settings. The judge LLM and thematic classification models we developed in Section \ref{sec:evaluation} are tuned specifically for our evaluation datasets and would require additional validation before application in other contexts \cite{wu-aji-2025-style, chen2024humans}. 

\textbf{Automated Evaluation} Some prior evaluations of minimally-adapted LLM use in the patient portal suggest that reduction in clinician time via LLM response drafting is minimal \cite{hu2025systematic, TaiSeale2024AIGeneratedDR, bootsma2025ai}. Our evaluation seeks to fill a critical research gap by automating the evaluation of how much a clinician would edit these responses, which we hope will enable progress towards better LLM alignment with individual clinicians and meaningful reduction in clinician workload. Our evaluation suggests that best-performing response drafting LLMs would reduce clinician edits by 25-26\%. This is a modest reduction, potentially due to the complexity of our data which covers real messages from general primary care and a wide range of medical topics and patient intents. Our focus on this automated evaluation limits us from performing in-depth qualitative analysis by clinicians and patients. While our hospital network is not an early adopter of LLM use in clinic which prohibits the use of our models for live patient messages, we hope to perform further studies with clinicians and patients in future work.

\textbf{Ethical Considerations} Real patient data used in our evaluations is highly sensitive, and extreme caution should be taken when using LLMs on real patient data to ensure patient privacy. We carefully design our evaluations to promote the responsible use of this data in our evaluation. Our data cleaning process ensures sensitive patients, e.g. patients under the age of 18, were not included in our final dataset. We host all real data on a secure server and perform all IPPM and SoCPPM experiments on this server. We only use proprietary LLMs on semi-synthetic data (SyPPM) which was created via completely de-identified patient charts and messages. 

\bibliography{custom}

\appendix
\newpage
\section{Dataset Details}
\label{app:dataset}

\subsection{Data Collection and Formatting}

As described in Section \ref{sec:data_overview}, the patient-clinician conversations used in our experiments are collected from a large academic hospital in the Eastern United States. These conversations are sourced from the hospital's electronic health record (EHR) portal messaging platform. 610k total messages are taken from the secure patient portal between 1/2020 - 9/2024. Our dataset includes messages from primary care, and thus includes a wide range of medical topics. We gather all patient-initiated messages which received a written clinician response to create 146k conversations, i.e. original patient message and response from a clinician. Our final data pool contains 10,105 unique patients, of which 64\% are female and 36\% are male, with ages ranging between 18-80. Each sample in our data pool consists of a patient message, a clinician response, and a summary of the patient's chart before the sending of the patient message. We designate 144k conversations from the data pool as training data, and we gather evaluation datasets from the remaining 2k conversations.

Details from throughout the EHR are summarized into four categories. First, the patient's age range and gender are given as \textbf{Demographics}. Next, the patient's active problems are listed under \textbf{Full Active Problem List}. The patient's recent encounters (with a maximum of 10 entries), including diagnoses, surgeries, visits, etc. are listed under \textbf{Recent Encounters}. Finally, a patient's outpatient medications are summarized in \textbf{Medications}. An example de-identified chart from SyPPM is provided in Figure \ref{app:fig:chart_summary}.

\begin{figure*}[t] 
\centering
\begin{fullwidthbox}
\subfile{prompts/ehr_summary}
\end{fullwidthbox}
\caption{Example de-identified EHR chart summary from our SyPPM patient message response drafting evaluation dataset}
\label{app:fig:chart_summary}
\end{figure*}

\subsection{Evaluation Dataset Details}
\label{app:dataset_eval_datasets}

Designating 144k training conversations, we gather evaluation datasets from the remaining 2k conversations. We create three evaluation sets, designed to evaluate LLM alignment with experts according to different standards of care. Each sample in each dataset is a tuple of strings $\{m, c, r\}$ consisting of a patient message $m$, a summary $c$ of the patient's EHR chart and a single clinician response $r$.

\textbf{IPPM} The Ideal Patient Portal Messaging (IPPM) dataset is created to evaluate LLMs in a setting where clinicians do not face the same resource constraints as in the real-world. In this evaluation dataset, ground-truth responses are written by a paid team of 4 expert primary care nurses who work daily in the patient portal, collected via REDCap surveys \cite{harris2009research}. In addition to giving ample time to write a full response to each message/EHR summary, experts were asked ``if you had unlimited time, what would be included in your response to this patient?'' To provoke quality responses, clinicians were given a separate text entry box for each of the themes derived in Section \ref{sec:data_themes}. For example, the \textit{Treatment Contingency Planning} text box included the prompt ``please outline a backup/red flag plan for the patient, if applicable.'' An example REDCap survey is given in Appendix \ref{app:redcap} for reproducibility. The IPPM dataset is comprised of 300 patient messages and corresponding EHR charts, with one expert clinician response per sample.

\textbf{SyPPM} As the other datasets use real patient data containing protected health information (PHI), they are not suitable for public release. We create the Synthetic Patient Portal Messaging (SyPPM) as a public benchmark to promote open-source research in clinician response drafting. We begin by taking 100 semi-synthetic patient portal messages which are created using a small number of de-identified patient portal messages in an in-context synthesis prompt \cite{gatto2025fuqg, gatto2024synth} and pair them with real de-identified patient EHR summaries. Ground-truth responses to each patient message are then provided by a primary care clinician, using the same theme-guided REDCap survey used for IPPM.

\textbf{SoCPPM} The Standards of Care Patient Portal Messaging (SoCPPM) dataset is created to evaluate LLMs in a practical setting, in which response drafts are compared with the clinician response which was sent via the secure portal in real time. This dataset is comprised of 300 patient messages and corresponding EHR summaries, where ground-truth responses are sourced from the patient portal. We evaluate LLM response drafts with respect to these real responses from the patient portal to study how LLM responses might perform in real-world settings, against the current standards of care in the patient portal. 

\section{Thematic Analysis Details}
\label{app:thematic_analysis}

We carefully derive elements of high-quality clinician responses to patient messages. Based on prior work, manual thematic analysis of real patient-clinician conversations, and consultation with expert primary care physicians, nurses, and triage nurses, we derive a set of ``themes'' which can be used to characterize the quality of clinician responses to patient messages \cite{braun2006using, sun2013messaging}. Below, we present our hybrid (top-down and bottom-up) approach to identify these themes.

As the quality of patient-clinician communication has a significant impact on patient health outcomes, characterizing quality response elements is important preliminary work for evaluating LLMs on the patient message response drafting task \cite{Stewart1995EffectivePC, doyle2013systematic}. Our goal is to derive themes that should occur in clinician responses to patient messages. We are interested in both empirically-derived themes, sourced from real patient-clinician conversations, as well as theoretically-derived themes, sourced from expert consultation and clinician communication theory \cite{Stewart1995EffectivePC, sakumoto2023digital}. Empirical themes are indicative of the current standards of care in patient portal communication, whereas theoretical themes may not be found in real-world clinician communication due to time, system, and resource constraints often experienced in asynchronous patient-clinician communication in the patient portal \cite{North2019ARA, Martinez2023PatientPM}. We therefore employ a hybrid top-down (theoretical), bottom-up (empirical) approach to identifying themes of quality clinician communication within the patient portal.

\subsection{Theoretical Response Themes}
We collaborate with a team of 11 clinicians to identify ``ideal'' clinician response themes to various patient messages. This iterative process involved 1-1 interviews with 2 primary care physicians and 9 primary care nurses, all of whom regularly interact with patients on the EHR portal from which our data pool (Appendix \ref{app:dataset}) is sourced. These interviews consisted of discussions based on open-ended questions, e.g. ``what are your primary goals when writing responses to patient messages in the patient portal?'' as well as discussions guided by examples of patient messages, e.g. ``what would you want to say to this patient?'' or "how would your response vary based on a <specific change> in the patient-initiated message?" Through these interviews, we derive an initial set of theoretical clinician response themes based on suggested best practices.

\subsection{Empirical Response Themes}
Using notes from these conversations as a backdrop, a team of three authors \footnote{Each team member is well-versed in health informatics and qualitative thematic analysis}, including a primary care physician, performed a comprehensive, iterative thematic analysis \cite{braun2006using} using a random sample of 100 patient messages, alongside a summary of the patient's electronic health record and the clinician's response. This process involved hand-labeling each sentence-length element of 25 clinician responses with a ``frame,'' then grouping those frames into ``themes,'' and repeating this process with new samples. In total we repeated this process four times. 

After performing the bottom-up thematic analysis, additional input from two primary care physicians guided the final, comprehensive list of eight clinician response themes comprised of 67 frames. Descriptions and examples of each response theme can be found in Table \ref{tab:themes}.

\section{EditJudge Framework Details}
\label{app:judge_model_details}

In Figure \ref{fig:eval_framework} we see an example of how the content-level and theme-level edit-F1 scores are calculated given a clinician response and an LLM response draft. In Algorithm \ref{alg:content_alg} we give the algorithm for counting expected matches $EM$, expected additions $EA$, and expected deletions $ED$ in an LLM-drafted response, in order to calculate content-level edit-F1 scores.


\begin{algorithm}
\caption{Counting expected matches $EM$, expected additions $EA$, and expected deletions $ED$ in an LLM-drafted response}
\label{alg:content_alg}
\begin{algorithmic}[1]
\Require $r_e$ (expert-written response), $r_d$ (LLM-drafted response)
\Ensure $EM$, $ED$, $EA$
\State Split $r_e$ into atomic elements (sentences)
\State Initialize $EM \gets 0$, $EA \gets 0$
\ForAll{sentence $s_e$ in $r_e$}
    \If{MATCH $s_e$ with content in $r_d$}
        \State $EM \gets EM + 1$
    \Else
        \State $EA \gets EA + 1$
    \EndIf
\EndFor
\State $r_d^- \gets$ Remove matching content from $r_d$
\State Split $r_d^-$ into sentences
\State $ED \gets$ number of sentences in $r_d^-$
\State \Return $EM$, $ED$, $EA$
\end{algorithmic}
\end{algorithm}

\subsection{Content-Matching Judge Model}

\begin{table*}
  \centering
  \footnotesize
  \setlength{\tabcolsep}{4pt}
  \begin{tabular}{llll}\hline
    \textbf{Clinician Sentence} &\textbf{LLM Draft (Excerpt)} &\textbf{Match Decision} &\textbf{Clinician Reasoning} \\\hline

    \makecell[lt]{Thank you for\\touching base -\\I'm sorry you\\have been\\experiencing these\\troubling symptoms.} &
    \makecell[lt]{Hi Sarah, Thank you for\\reaching out and bringing\\this to my attention.\\Postmenopausal bleeding can\\indeed be a sign of something\\more serious, so it's important\\that we investigate further.} &
    \makecell[lt]{Thank you for reaching out\\and bringing this to\\my attention.} &
    \makecell[lt]{Both sentences thank the\\patient for being proactive\\and express sympathy about\\their symptoms. The clinician\\would not have to rewrite\\this component in order to\\achieve the same meaning.} \\\hline
    
    \makecell[lt]{Please let us know\\if you have\\any issues with\\getting the\\medication.} &
    \makecell[lt]{Thanks for reaching out to me\\about your decision to switch.\\I would recommend a\\different medication.\\Please let me know if you have\\questions or concerns.\\We can discuss this further\\in your upcoming appointment.} &
    \makecell[lt]{NO MATCH} &
    \makecell[lt]{While issues with getting\\the medication may be\\classified as a concern,\\the draft is not specific\\enough and the clinician\\would have to rewrite.} \\\hline
    
    \makecell[lt]{Have you eaten\\anything out of\\the ordinary\\for you?} &
    \makecell[lt]{Have you experienced any\\other symptoms, such as\\stomach pain, bloating,\\or changes in bowel movements?\\Are you taking any new\\medications or supplements\\that could be causing the nausea?\\Have you recently changed\\your diet or experienced\\any significant stress?} &
    \makecell[lt]{Have you recently changed\\your diet or experienced\\any significant stress?} &
    \makecell[lt]{Both symptom-related follow-up\\questions ask the patient\\about recent diet changes,\\and the clinician would not\\have to rewrite the drafted\\sentence in order to achieve\\the same meaning.} \\\hline
    
  \end{tabular}
  \caption{Selected examples from the content-level editJudge evaluation dataset. The editJudge model is given the LLM draft (an excerpt from each is shown in this table to preserve space) and a sentenec from a clinician-written response, and is tasked with outputting either the matching content from the LLM draft, or the string ``NO MATCH''. We show two matching decisions, one from the empathetic communication theme and another from the symptom-related follow-up question theme, as well as a close non-match from the contingency planning theme.}
  \label{app:tab:matches}
\end{table*}
\begin{table*}
  \centering
  \begin{tabular}{|l|l|cccc|}\hline
    \textbf{Model} &\textbf{Type} &\textbf{Avg Agr} &\textbf{Avg Non-Match} &\textbf{Avg Match} &\textbf{\% Match} \\\hline
    Qwen2.5-7B-Instruct &0-Shot &0.74 &1.00 &0.07 &0.07 \\
    Llama-3-8B-Instruct &0-Shot &0.17 &0.11 &0.32 &0.50 \\\hline
    Qwen2.5-7B-Instruct &5-Shot &0.71 &0.93 &0.14 &0.14 \\
    Llama-3-8B-Instruct &5-Shot &0.63 &0.88 &0.00 &0.00 \\\hline
    Qwen2.5-3B &SFT &0.76 &0.97 &0.21 &0.21 \\
    Qwen2.5-3B-Instruct &SFT &0.80 &0.94 &0.43 &0.50 \\
    Llama-3.2-3B-Instruct &SFT &0.85 &1.00 &0.46 &0.57 \\
    Qwen2.5-7B &SFT &0.87 &0.97 &0.61 &0.71 \\
    Qwen2.5-7B-Instruct &SFT &0.89 &0.97 &0.68 &0.71 \\
    Llama-3-8B-Instruct &SFT &\textbf{0.96} &\textbf{1.00} &\textbf{0.84} &\textbf{0.92} \\\hline
  \end{tabular}
  \caption{EditJudge model performance across different configurations. We find that SFT is superior to either 0-shot or 5-shot editJudge models. We find that the best model, the fine-tuned instruction-tuned Llama3-8B model, achieves 96\% agreement with clinician-guided author annotations. 84\% of the \textit{matching} author annotations were exactly matched by this judge model, and 92\% of match decisions contained at least some overlap.}
  \label{app:tab:match_results}
\end{table*}

Here we describe the process used to fine-tune the content-level editJudge model used in Algorithm \ref{alg:content_alg} to calculate content-level edit-F1. First, three authors hand-label 450 training samples and 50 evaluation samples. Each sample input is a response draft written by the Aloe-8B \cite{gururajan2024aloe} 0-shot model, along with a sentence drawn from an expert-written response to a sample from the publicly-available SyPPM evaluation dataset. The annotators either wrote ``NO MATCH'' if there was no matching content from the response draft, or copy/pasted the matching content from the response draft if applicable. The prompt to identify matches was ``if the expert clinician would not have to rewrite this content in order to achieve the same meaning as their given sentence, this is matching content.'' Author annotators were asked to flag all samples about which they were unsure or which required clinical expertise, and two expert clinicians were consulted on these samples to make a final decision.

This matching decision is not always straightforward. For example, in Figure \ref{fig:eval_framework} we see that the clinician-written sentence ``I'm sorry to hear about your new symptoms'' matches with the LLM-drafted sentence ``I'm sorry you've been feeling nauseous.'' While expert clinicians in our evaluation agreed that they would not need to rewrite this LLM-drafted sentence, in order to achieve the same meaning as the clinician-written sentence, this is not always trivial and may vary from clinician to clinician.  Examples of clinician-verified matches and non-matches from our training samples can be found in Table \ref{app:tab:matches}. 

Given a sentence $s_e$ from an expert-written response $r_e$ and an LLM-drafted response $r_d$ the content-level editJudge model was tasked with outputting either the matching content from the LLM draft $s_d$, or the string ``NO MATCH''. Since the matching content $s_d$ is later removed from $r_d$ to identify expected deletions $ED$, the output of the judge model $\hat{s}_d$ must match verbatim to the matching content in the draft $s_d$ in order to remove $s_d$ in Algorithm \ref{alg:content_alg}. We therefore evaluate the editJudge model by identifying whether it outputs exactly-matching content $s_d$ identified by the annotators. We first identify whether the editJudge model correctly makes the matching decision (either by outputting ``NO MATCH'' or some substring $\hat{s}_d$ from the LLM draft $r_d$), and call this \textbf{agreement}, i.e. the proportion of evaluation samples on which the judge model makes the correct matching decision. We further score the editJudge model by identifying \textbf{non-match agreement}, i.e. the proportion of non-matches correct identified by the judge model, and \textbf{match agreement}, the proportion of annotated which are exactly matched by the editJudge model outputs. To get a granular estimate of judge model outputs, we also score \textbf{match overlap}, i.e. the proportion of evaluation responses in which editJudge model output $\hat{s}_d$ and annotated matching content $s_d$ overlap. We evaluated 6 judge models, testing 0-shot, 5-shot, and supervised fine-tuning adaptation strategies for this content-level matching task. 

We see content-level judge results in Table \ref{app:tab:match_results}. In general, SFT is far superior to either 0-shot or 5-shot judge models. We find that the best model, the instruction-tuned Llama3-8B model \cite{llama3modelcard} fine-tuned on the $450$ training samples, achieves 96\% agreement with clinician-guided author annotations. 84\% of the \textit{matching} author annotations were exactly matched by this judge model, meaning the exact correct content would be removed from the LLM draft $r_d$ to identify exact expected deletions $ED$, and 92\% of match decisions contained at least some overlap. 

\subsection{Sentence Theme Classification Model}

We now similarly describe the fine-tuning the sentence-level theme classification model, used to calculate the theme-level edit-F1 score described in Section \ref{sec:data_themes}. First, one author hand-labeled 175 training samples and 50 evaluation samples. Each sample was a sentence-length string taken from responses to SyPPM samples generated by the Aloe-8B \cite{gururajan2024aloe} 0-shot. Consulting with two expert clinicians, each sample was assigned a single theme label, including the 8 themes and an ``Other'' label, to set up a 9-class classification task. Example sentences from each theme can be found in Table \ref{tab:themes}.

\begin{table}
  \centering
  \begin{tabular}{|l|c|}\hline
    \textbf{Theme} &\textbf{F1} \\\hline
    \makecell[l]{Empathetic Communication} &0.94 \\\hline
    \makecell[l]{Symptom-Related\\Follow-Up Questions} &1.00 \\\hline
    \makecell[l]{Medication-Related\\Follow-Up Questions} &0.67 \\\hline
    \makecell[l]{Medical Assessment Explanation} &0.67 \\\hline
    \makecell[l]{Medical Planning Instruction} &0.71 \\\hline
    \makecell[l]{Logistics: Scheduling,\\Billing, Operations} &0.82 \\\hline
    Care Coordination &0.80 \\\hline
    \makecell[l]{Contingency Planning} &0.67 \\\hline
    Other &1.00 \\\hline
    \textit{Micro Avg} &\textit{0.82} \\
    \hline
  \end{tabular}
  \caption{Sentence classification model results. Using a fine-tuned Llama3-8B model \cite{llama3modelcard}, we report class-wise performance and micro average F1. We see that the sentence classification model performs well overall, with a micro average F1 of 0.82, and that it predicts all individual classes competently (> 0.67 F1).}
  \label{app:tab:theme_results}
\end{table}

Following the results of the content-level editJudge training, we choose to fine-tune a Llama3-8B model \cite{llama3modelcard} to perform the sentence classification, where the task is to output the class label (e.g. ``Symptom-Related Follow-Up Question'') given the response sentence. Class-wise performance and micro average F1 of this sentence classification model are reported in Table \ref{app:tab:theme_results}. We see that the sentence classification model performs well overall, with a micro average F1 of 0.82, and that it predicts all individual classes competently (> 0.67 F1). We note that this task is subjective on some level, given that theme classes are not necessarily disjoint. For example, there are valid reasons to argue that a question such as ``have you noticed any diarrhea while on your amoxicillin?'' could be both a symptom- and medication-related follow-up question. However, we enforce a single-class label for simplicity in our evaluations.

\section{LLM Adaptation Details}
\label{app:adaptation_details}

As described in Section \ref{sec:exp_setup_adapts}, here we provide details for the supervised fine-tuning (SFT) and thematic agentic direct preference optimization for learning enhancement (TADPOLE) LLM adaptation strategies which we use in our evaluation in Section \ref{sec:exp_results}. Prompts for the 0-shot and thematic adaptations can be found in Appendix \ref{app:prompts}. Further details for the RAG, SFT, and TADPOLE adaptations can be found below. 

\subsection{RAG Details}
\label{app:adaptation_rag}

Using the training dataset (144k) as a RAG database, we encode patient messages and EHR summaries using S-BERT\footnote{all-MiniLM-L6-v2} \cite{reimers2019sentence}, and include the 5 most similar message + EHR strings, along with their real clinician responses in the prompt to guide the LLM, alongside the instruction from the 0-shot prompt.

\subsection{SFT Details}
\label{app:adaptation_sft}

We perform supervised fine-tuning using all 144k training messages. The LLM is trained to output the clinician response $r$, given the patient message $m$ and a summary of the patient's EHR $c$ contextualized with the 0-shot prompt (see Appendix \ref{app:prompts} for this prompt).

Each time a model is fine-tuned, both for the SFT models in Section \ref{sec:exp_setup_adapts} and for the fine-tuned judge models in Section \ref{sec:evaluation}, we train for 1 epoch using a batch size of 4 on a single Nvidia A40 GPU (48GB RAM). We train using low-rank adaptation (LoRA) \cite{hu2022lora} for efficiency, which has shown to be a performant fine-tuning strategy \cite{shulman2025lora, zhao2024lora}. We use LoRA with rank 8 and an alpha scaling factor of 16. We use the AdamW optimizer with weight decay of 0.01, linear learning rate scheduler with warmup over 10\% of the training steps, and gradient clipping at a norm of 1.0. We apply mixed precision training using float16 to optimize memory usage and training speed.

\subsection{TADPOLE Details}
\label{app:adaptation_tadpole}

For each theme, TADPOLE takes a a base response $r$ and creates both an ``enhanced'' response $r^+$ and ``corrupted'' response $r^-$ by either adding or removing thematic content from the response. First, we take 8k training samples and generate base responses using the fine-tuned (SFT) model. For \textit{enhancing} a response $r$ with content from a given theme $t$, we use a response enhancing agent to get an enhanced response $r_t^+$. Each thematic enhancement agent is a simple 3-shot prompt. For \textit{corrupting} a response $r$ with content from a given theme $t$, we use a standard corruption agent contextualized with the theme $t$ to obtain a corrupted response $r_t^-$. Enhancement prompts and the corruption prompts are developed for and passed to the Qwen2.5-32B-Instruct\footnote{Qwen/Qwen2.5-32B-Instruct} \cite{qwen25} model. We obtain 1k enhanced responses for each theme and 1k corrupted responses for each theme for a total of 8k enhanced, base, and corrupted response $\{r^+, r, r^-\}$ tuples. 

Following \citet{lialfa}, we test several preference pair creation strategies using these tuples. \textbf{Enhanced} pairs $\{r^+, r\}$ use enhanced responses and base responses as chosen and rejected responses, respectively. \textbf{Corrupted} pairs $\{r, r^-\}$ choose base responses over corrupted responses. \textbf{Hard-Corrupted} pairs $\{r^+, r^-\}$ choose enhanced responses over corrupted responses. We also investigate a \textbf{Blend} which contains an even amount of all three pairs. We perform DPO \cite{rafailov2023direct} on the fine-tuned model using 8k TADPOLE preference pairs. We perform DPO on the SFT model using a beta of 0.01. Similarly with SFT, we perform DPO by training for 1 epoch using a batch size of 1 on a single Nvidia A40 GPU (48GB RAM). We apply mixed precision training using float16 to optimize memory usage and training speed.


\begin{table*}
  \centering
  \begin{tabular}{|c|ccc|ccc|}\hline
    \textbf{} &\multicolumn{3}{c}{\textbf{Content-Level}} &\multicolumn{3}{|c|}{\textbf{Theme-Level}} \\\hline
    \textbf{Pairs} &\textbf{Pr} &\textbf{Re} &\textbf{Edit-F1} &\textbf{Pr} &\textbf{Re} &\textbf{Edit-F1} \\\hline
    Blend &0.13 &\textbf{0.19} &\textbf{0.14} &0.53 &\textbf{0.65} &0.58\\
    Enhanced &0.09 &0.14 &0.10 &0.45 &0.62 &0.52 \\
    Corrupted &0.13 &0.16 &0.12 &\textbf{0.60} &0.62 &\textbf{0.61} \\
    Hard-Corrupted &\textbf{0.13} &0.18 &\textbf{0.14} &0.54 &\textbf{0.65} &0.59 \\\hline
    \textit{IAP} &\textit{0.26} &\textit{0.25} &\textit{0.24} &\textit{0.61} &\textit{0.63} &\textit{0.62} \\
    \hline
    \end{tabular}
  \caption{Content-level and theme-level edit-F1 scores for varying TADPOLE preference pair creation strategies on the IPPM dataset. The hard-corrupted strategy achieves best performance at the content-level, as well as overall when weighting evenly between content- and theme-level edit-F1 scores.}
  \label{app:tab:tadpole_content}
\end{table*}

We report average content-level and theme-level edit-F1 scores on IPPM for each TADPOLE strategy in Table \ref{app:tab:tadpole_content}. The hard-corrupted strategy achieves best performance at the content-level, as well as overall when weighting evenly between content- and theme-level edit-F1 scores. Hence we report the results of the models trained on hard-corrupted pairs in Section \ref{sec:exp_results}.

\section{Measures of Inter-Clinician Variation}

\subsection{Inter-Annotator Agreement}
\label{app:iaa}
Clinician responses to patient messages may vary based on experience factors (e.g. role, years of experience, specialty), personality factors (e.g. writing style), and interpersonal factors (e.g. relationship with the patient). Table \ref{app:tab:multiresponse} gives examples of different clinician responses to the same patient message within the SyPPM dataset. 

As noted in Section \ref{sec:eval_iaa}, we gather 3 expert responses to 40 samples from the SyPPM dataset. Of the 3 experts, 1 is a primary care physician with 15+ years of experience and 2 are primary care nurses, each with 5+ years of experience. In Section \ref{sec:eval_iaa} we describe how we might use multiple responses to understand inter-annotator predictability (IAP). Here we describe three measures of inter-annotator agreement (IAA), using these same samples.

\begin{table*}
  \centering
    \begin{tabular}{|c|cccccccc|}\hline
        \textbf{IAA Measure} &\textbf{Emp} &\textbf{Sym Q} &\textbf{Med Q} &\textbf{Asse} &\textbf{Plan} &\textbf{Log} &\textbf{Coord} &\textbf{Cont} \\\hline
        Strict Inclusion &0.53 &0.53 &0.20 &0.03 &0.00 &0.57 &0.00 &0.00 \\
        Strict Exclusion &0.00 &0.00 &0.00 &0.33 &0.93 &0.00 &0.33 &0.47 \\\hline
        Strict Agreement &0.53 &0.53 &0.20 &0.36 &0.93 &0.57 &0.33 &0.47 \\\hline
    \end{tabular}
  \caption{Inter-annotator agreement (IAA) measured at the theme-level by identifying cases when all three annotators either included (strict inclusion) or excluded (strict exclusion) each theme in their response. We find that some themes are unanimously found in all clinician responses to most (> 50\%) patient messages. Interestingly, we also find that the medical treatment theme is almost never found in any clinician response to most patient messages (< 7\%). This speaks to the reluctance of these clinicians to treat patients via the portal, instead favoring information seeking (e.g. follow-up questions) responses.}
  \label{app:tab:iaa_strict}
\end{table*}

We are interested in measuring how similarly clinicians would respond to the same patient message in the same conditions. We start by identifying, for each theme, the proportion of patient messages to which all three annotator responses either included that theme (\textbf{strict inclusion}), or did not include that theme (\textbf{strict exclusion}). Taken together (\textbf{strict agreement}), we can estimate the extent to which each response theme is clinician-independent. 

These theme-level IAA measurements can be found in Table \ref{app:tab:iaa_strict}. We find that themes such as empathetic communication, symptom-related follow-up questions, and logistical information are unanimously found in all clinician responses to most (> 50\%) patient messages in SyPPM. Interestingly, we also find that the medical treatment theme is almost never found in any clinician response to most patient messages (< 7\%). This speaks to the reluctance of these clinicians to treat patients via the portal, instead favoring information seeking (e.g. follow-up questions) responses.

\begin{table}
  \centering
  \begin{tabular}{|c|ccc|}\hline
    \textbf{Clinician} &\textbf{A} &\textbf{B} &\textbf{C} \\\hline
    \textbf{A} &1.00 &0.51 &0.59 \\
    \textbf{B} &0.51 &1.00 &0.45 \\
    \textbf{C} &0.59 &0.45 &1.00 \\
    \hline
  \end{tabular}
  \caption{Inter-annotator agreement measured at the content-level between clinician pairs using cosine similarity. We find that agreement between clinician pairs varies substantially, with some  (clinicians A and C) more aligned than others (clinicians B and C).}
  \label{app:tab:iaa_cosine}
\end{table}

For a simpler measure of IAA, we also measure the average pairwise cosine similarity of each clinician's responses, comparing each pair of clinicians in Table \ref{app:tab:iaa_cosine}. We find that agreement between clinician pairs varies substantially, with some (clinicians A and C, 0.59) more aligned than others (clinicians B and C, 0.45).

\subsection{Inter-Annotator Predictability}

We calculate IAP using both content-level and theme-level edit-F1 scores to enable direct comparison to our model results in Section \ref{sec:exp_results}. To estimate the amount of agreement between two expert clinicians in our evaluation framework, we assign the first clinician the role of expert and the second the role of drafting responses. Treating the first clinician's response as the expert response $r_e$ and the second's response as the response draft $r_d$, we calculate content-level and theme-level edit-F1 scores using the editJudge described in Section \ref{sec:evaluation}. Assigning each ordered pair ($N=6$) of expert responses as ground-truth responses and response drafts, we compare $6 \times 40 = 240$ total responses, and take the average results.  Tables \ref{tab:combined_results}, \ref{app:tab:content_classwise}, and \ref{app:tab:theme_classwise} give IAP estimates for content-level and theme-level edit-F1 scores, class average content-level recall scores, and class average theme-level edit-F1 scores, respectively.

\begin{table*}
  \centering
  \footnotesize
  \setlength{\tabcolsep}{4pt}
  \begin{tabular}{lll}\hline
    \textbf{Patient Message} &\textbf{Clinician A Response} &\textbf{Clinician B Response} \\\hline
    \makecell[lt]{I'm not feeling quite\\myself lately. I've been\\experiencing some weakness\\that's making everyday\\activities a bit more\\challenging. I was wondering\\if we could touch base\\about what might be\\causing this?} &
    \makecell[lt]{I'm sorry you have been\\experiencing these troubling\\symptoms. Could you describe\\where you experience the\\weakness? How does it impede\\your daily activities? Does it\\come and go? How long has\\this been going on? Do you\\have any other symptoms such\\as dizziness or lightheadedness?\\Have you checked your blood\\pressure at home? Have you had\\any changes to your medications\\recently? Please call the office\\to schedule an appointment for\\urgent evaluation. If your symptoms\\worsen acutely, including any\\dizziness or lightheadedness, or\\syncopal episodes (fainting), you\\should call 911 and be seen\\emergently in the ER.} &
    \makecell[lt]{Sorry to hear you aren't feeling\\well. Are you having any other\\symptoms? How long have these\\symptoms been going on? Have\\you ever had symptoms like this\\before? Are you having any nausea,\\vomiting, diarrhea, or constipation?\\Are you having any fevers? Are you\\losing weight without trying? Are\\you having any blood in bowel\\movements? Are you having\\abdominal pain? Have you noticed\\any particular foods that trigger\\the symptoms? Have you started\\any new medications or supplements?\\Have you recently changed dosing\\or timing of medications you take?\\Have you tried any medications\\that have helped? Please give us\\a call to schedule an appointment.\\You should be seen in the ED if\\you have worsening or sudden\\abdominal pain, severe vomiting,\\dizziness, chest pain, or shortness\\of breath.} \\\hline
    \makecell[lt]{I'm having a pretty rough\\time with my seasonal\\allergies right now. My eyes\\are itchy, I'm congested, and\\I just can't seem to stop\\sneezing. I've been using\\some over-the-counter meds,\\but they're not really giving\\me the relief I need. I was\\wondering if you could\\recommend something a bit\\stronger or if I should come\\in for an appointment.} &
    \makecell[lt]{I'm sorry you have been\\experiencing these troubling\\symptoms. Which medications\\have you tried, and what has\\helped you in the past?} &
    \makecell[lt]{Are you having any other\\symptoms? Are you having\\any fevers? Are you having\\any shortness of breath?\\Have you started any new\\medications or supplements?\\Have you recently changed\\dosing or timing of medications\\you take? Have you tried any\\medications that have helped?\\Please give us a call to\\schedule an appointment. Give\\our triage nurses a call if\\your symptoms are worsening.} \\\hline
    \makecell[lt]{I've been dealing with itchy\\eyes for weeks now, and I'm\\guessing it's just my allergies\\acting up again. I was wondering\\if I could get your thoughts\\on it - should I just stick with\\my usual meds or is there\\something else I can try?} &
    \makecell[lt]{I'm sorry that you have been\\experiencing these troubling\\symptoms. Have you been having\\any other symptoms? Have you\\had any recent changes in your\\medications? Have you tried\\anything that may have helped\\alleviate your symptoms? If your\\symptoms are persisting on your\\usual allergy medications, or\\symptoms are worsening, please\\call the office to schedule an\\appointment.} &
    \makecell[lt]{Thanks for checking in. Are you\\having any other symptoms? How\\long have these symptoms been\\going on? Have you ever had\\symptoms like this before? Have\\you started any new medications\\or supplements? Have you recently\\changed dosing or timing of\\medications you take? Have you\\tried any medications that have\\helped? Please call to schedule\\an appointment. You should be\\seen in the ED if you have\\worsening or sudden shortness\\of breath, vision changes, or\\chest pain.} \\\hline
  \end{tabular}
  \caption{Examples of different clinician responses to the same patient message within the SyPPM dataset. We collect responses from three separate annotators to 40 messages within the SyPPM dataset, and show selected examples from two annotators here.}
  \label{app:tab:multiresponse}
\end{table*}


\section{Additional Results}
\label{app:additional_results}






\subsection{SocPPM Results}
\label{app:additional_results:socppm}



\begin{table*}[h]
\centering
    \begin{tabular}{|c|c|ccc|ccc|}
    \hline
    \multicolumn{2}{|c|}{} & \multicolumn{3}{c}{\textbf{Content-Level}} & \multicolumn{3}{|c|}{\textbf{Theme-Level}} \\ \hline
    Dataset & Model & Precision & Recall & Edit-F1 & Precision & Recall & Edit-F1 \\ \hline
    \multirow{5}{*}{\textbf{SoCPPM}}
    &0-Shot & 0.06\ps0.01 & 0.29\ps0.08 & 0.10\ps0.01 &0.48\ps0.01 &0.83\ps0.08 &0.61\ps0.01 \\ 
    &Theme & 0.06\ps0.00 & 0.32\ps0.11 & 0.09\ps0.01 &0.44\ps0.00 &\textbf{0.85\ps0.11} &0.58\ps0.01 \\ 
    &RAG & 0.11\ps0.03 & \textbf{0.33\ps0.18} & 0.14\ps0.01 &0.49\ps0.03 &0.75\ps0.18 &0.59\ps0.01 \\ 
    &SFT & \textbf{0.15\ps0.01} & 0.18\ps0.00 & \textbf{0.15\ps0.01} &\textbf{0.63\ps0.01} &0.62\ps0.00 &\textbf{0.62\ps0.01} \\ 
    &TADPOLE & 0.12\ps0.01 & 0.19\ps0.01 & 0.14\ps0.01 &0.51\ps0.01 &0.69\ps0.01 &0.59\ps0.01 \\ \hline
    \multicolumn{2}{|c|}{\textit{IAP}} & \textit{0.26} & \textit{0.25} & \textit{0.24} &\textit{0.61} &\textit{0.63} &\textit{0.62} \\ \hline
    \end{tabular}
    \caption{Edit-F1 scores for LLM adaptations on the SoCPPM patient message response drafting dataset. Each model adaptation is performed on three underlying LLMs, we report scores as average\ps standard deviation. We report content-level precision, recall, and edit-F1 (Section \ref{sec:exp_setup:content}), as well as theme-level precision, recall, and edit-F1 (Section \ref{sec:exp_setup:themes}). We report content-level inter-annotator predictability (IAP), comparing LLM performance and expert human alignment.}
    \label{app:tab:combined_soc}
\end{table*}

The SoCPPM dataset is created to evaluate LLMs in a practical setting, in which response drafts are compared with the clinician response which was sent via the secure portal in real time. In some ways this is a less-ideal form of the patient message response drafting task, because real-time clinician responses tend to contain a high degree of variation which is challenging to filter automatically. For example, real-time clinician responses frequently contain standardized responses (``dot phrases'') which offer commonly-repeated instructions, e.g. ``please call the COVID-19 hotline if you are experiencing any of the following symptoms...'' \cite{baltaro2022patient}. Additionally, real-time responses are written under more duress due to workforce constraints and growing use of the patient portal \cite{Budd2023BurnoutRT, Underdahl2024PhysicianBE, Martinez2023PatientPM, yan2021exploring}.

In Table \ref{app:tab:combined_soc} we report the content-level and theme-level precision, recall and edit-F1 scores for adapted LLMs on the SoCPPM dataset. We find that LLMs in general perform more poorly on this dataset than the ideal IPPM and SyPPM datasets. The best-performing model adaptation on SyPPM (TADPOLE) achieves 0.20 content-level edit-F1 scores on SyPPM (see Table \ref{tab:combined_results}). The best-performing model adaptation on SoCPPM (SFT) achieves only 0.15 content-level edit-F1 on SoCPPM (Table \ref{app:tab:combined_soc}). We hypothesize that this is because the SoCPPM dataset represents a version of the patient message response drafting task that is both more challenging, due to the existence of situated knowledge scattered throughout the EHR system that is unknowable for the response drafting LLM, and less ideal, given that frequently clinician responses in practical settings can be messy and often sent under time pressure \cite{Budd2023BurnoutRT, Underdahl2024PhysicianBE, Martinez2023PatientPM, yan2021exploring}.

We also note that SFT outperforms TADPOLE on the SoCPPM dataset, with SFT achieving 0.15 and 0.62 content- and theme-level edit-F1 scores, respectively, and TADPOLE achieving only 0.14 and 0.59 (Table \ref{app:tab:combined_soc}). As TADPOLE adaptation uses thematic preference pairs to further fine-tune SFT models, we hypothesize that the themes used to generate these preference pairs are less suitable for the lower-quality, higher-variation responses found in real-time clinician responses.


\subsection{Class-Average Edit-F1 Scores}
\label{app:additional_results_class_average}

\begin{table*}[h]
\footnotesize
\centering
\begin{tabular}{|c|l|c|c|c|c|c|c|c|c|}
\hline
\textbf{Dataset} & \textbf{Model} & \textbf{Emp} & \textbf{SymQ} & \textbf{MedQ} & \textbf{Assess} & \textbf{Plan} & \textbf{Logis} & \textbf{Coord} & \textbf{Cont} \\ \hline
\multirow{6}{*}{\textbf{SoCPPM}} &\textit{Proportion} &\textit{0.81} &\textit{0.05} &\textit{0.02} &\textit{0.38} &\textit{0.27} &\textit{0.42} &\textit{0.58} &\textit{0.03} \\ \cline{2-10}
&0-Shot & 0.29\ps0.02 & 0.07\ps0.00 & 0.07\ps0.00 & 0.16\ps0.01 & 0.23\ps0.01 & 0.21\ps0.02 & 0.24\ps0.01 & 0.21\ps0.02 \\ 
&Theme & 0.30\ps0.02 & 0.07\ps0.01 & 0.07\ps0.00 & \textbf{0.16\ps0.00} & 0.23\ps0.00 & 0.23\ps0.03 & \textbf{0.25\ps0.00} & 0.24\ps0.04 \\ 
&RAG & 0.30\ps0.02 & 0.07\ps0.00 & 0.07\ps0.00 & 0.16\ps0.01 & \textbf{0.25\ps0.03} & 0.22\ps0.03 & 0.25\ps0.01 & 0.23\ps0.04 \\ 
&SFT & \textbf{0.30\ps0.01} & \textbf{0.08\ps0.00} & 0.07\ps0.00 & 0.15\ps0.00 & 0.21\ps0.01 & \textbf{0.24\ps0.00} & 0.24\ps0.00 & \textbf{0.27\ps0.00} \\ 
&TADPOLE & 0.30\ps0.02 & \textbf{0.08\ps0.00} & 0.07\ps0.00 & 0.15\ps0.01 & 0.23\ps0.03 & 0.23\ps0.02 & 0.24\ps0.00 & 0.25\ps0.04 \\ \hline
\multirow{6}{*}{\textbf{IPPM}} &\textit{Proportion} &\textit{0.76} &\textit{0.23} &\textit{0.09} &\textit{0.31} &\textit{0.24} &\textit{0.51} &\textit{0.67} &\textit{0.07} \\ \cline{2-10}
&0-Shot & 0.28\ps0.02 & 0.07\ps0.00 & \textbf{0.07\ps0.00} & 0.15\ps0.01 & 0.23\ps0.02 & 0.21\ps0.03 & 0.24\ps0.00 & 0.22\ps0.04 \\ 
&Theme & 0.29\ps0.02 & 0.07\ps0.02 & 0.06\ps0.01 & \textbf{0.15\ps0.00} & 0.28\ps0.09 & 0.23\ps0.02 & 0.25\ps0.01 & 0.22\ps0.07 \\ 
&RAG & 0.24\ps0.06 & 0.04\ps0.03 & 0.04\ps0.03 & \textbf{0.15\ps0.00} & \textbf{0.33\ps0.09} & 0.19\ps0.06 & \textbf{0.26\ps0.01} & 0.20\ps0.06 \\ 
&SFT & \textbf{0.30\ps0.00} & \textbf{0.08\ps0.00} & \textbf{0.07\ps0.00} & \textbf{0.15\ps0.00} & 0.21\ps0.00 & \textbf{0.24\ps0.00} & 0.24\ps0.00 &\textbf{ 0.27\ps0.00} \\ 
&TADPOLE & \textbf{0.30\ps0.00} & \textbf{0.08\ps0.00} & \textbf{0.07\ps0.00} & 0.15\ps0.01 & 0.21\ps0.02 & 0.24\ps0.01 & 0.24\ps0.01 & 0.26\ps0.02 \\ \hline
\multirow{8}{*}{\textbf{SyPPM}} &\textit{Proportion} &\textit{0.99} &\textit{0.79} &\textit{0.79} &\textit{0.33} &\textit{0.05} &\textit{0.75} &\textit{0.11} &\textit{0.56} \\ \cline{2-10}
&0-Shot & 0.28\ps0.03 & 0.06\ps0.02 & 0.07\ps0.01 & 0.15\ps0.00 & \textbf{0.27\ps0.05} & 0.22\ps0.01 & 0.24\ps0.00 & 0.22\ps0.02 \\ 
&Theme & 0.30\ps0.02 & \textbf{0.08\ps0.00} & \textbf{0.08\ps0.00} & \textbf{0.16\ps0.01} & 0.24\ps0.01 & 0.24\ps0.03 & \textbf{0.25\ps0.01} & 0.25\ps0.05 \\ 
&RAG & \textbf{0.31\ps0.01} & \textbf{0.08\ps0.00} & 0.07\ps0.00 & \textbf{0.16\ps0.01} & 0.23\ps0.02 & \textbf{0.25\ps0.01} & 0.24\ps0.01 & 0.26\ps0.01 \\
&SFT & 0.29\ps0.03 & 0.06\ps0.02 & 0.06\ps0.02 & 0.15\ps0.01 & 0.27\ps0.06 & 0.24\ps0.01 & \textbf{0.25\ps0.01} & 0.22\ps0.05 \\ 
&TADPOLE & 0.30\ps0.00 & \textbf{0.08\ps0.00} & 0.07\ps0.00 & 0.15\ps0.01 & 0.21\ps0.01 & 0.24\ps0.01 & 0.24\ps0.01 & \textbf{0.27\ps0.00} \\
&\textit{Gemini} &\textit{0.30} &\textit{0.07} &\textit{0.07} &\textit{0.16} &\textit{0.23} &\textit{0.24} &\textit{0.24} &\textit{0.24} \\ \hline
\multicolumn{2}{|c|}{\textit{IAP}} &\textit{0.30} &\textit{0.07} &\textit{0.07} &\textit{0.16} &\textit{0.23} &\textit{0.24} &\textit{0.24} &\textit{0.24} \\ \hline
\end{tabular}
\caption{Class average content-level recall scores for adapted LLMs. Each model adaptation is performed on three underlying LLMs, we report average results \ps standard deviation. We report micro average recall scores for each theme class. We also report the proportion of responses which contain each theme in each dataset. We include SyPPM results of the best commercial model (Gemini with theme prompting) for comparison. Finally, we report theme-level IAP, comparing LLM performance and expert human alignment at the theme level.}
\label{app:tab:content_classwise}
\end{table*}

\begin{table*}[h]
    \footnotesize
    \centering
    \begin{tabular}{|c|l|c|c|c|c|c|c|c|c|}
    \hline
    \textbf{Dataset} & \textbf{Model} & \textbf{Emp} & \textbf{SymQ} & \textbf{MedQ} & \textbf{Assess} & \textbf{Plan} & \textbf{Logis} & \textbf{Coord} & \textbf{Cont} \\ \hline
    \multirow{6}{*}{\textbf{SoCPPM}} &\textit{Proportion} &\textit{0.81} &\textit{0.05} &\textit{0.02} &\textit{0.38} &\textit{0.27} &\textit{0.42} &\textit{0.58} &\textit{0.03} \\ \cline{2-10}
    &0-Shot & 0.88\ps0.02 & 0.12\ps0.06 & 0.08\ps0.07 & \textbf{0.57\ps0.02} & 0.46\ps0.02 & \textbf{0.58\ps0.03} & 0.68\ps0.02 & \textbf{0.12\ps0.07} \\ 
    &Theme & 0.88\ps0.02 & 0.13\ps0.05 & \textbf{0.12\ps0.05} & 0.55\ps0.00 & 0.44\ps0.02 & \textbf{0.58\ps0.03} & 0.69\ps0.02 & 0.09\ps0.01 \\ 
    &RAG & 0.82\ps0.05 & 0.12\ps0.05 & 0.07\ps0.06 & 0.55\ps0.01 & \textbf{0.47\ps0.01} & 0.57\ps0.03 & \textbf{0.70\ps0.03} & 0.11\ps0.09 \\ 
    &SFT & 0.88\ps0.01 & 0.10\ps0.16 & 0.00\ps0.00 & 0.42\ps0.02 & 0.36\ps0.07 & 0.51\ps0.04 & 0.64\ps0.02 & 0.09\ps0.08 \\ 
    &TADPOLE & \textbf{0.89\ps0.00} & \textbf{0.18\ps0.04} & 0.10\ps0.04 & 0.45\ps0.02 & 0.35\ps0.07 & 0.53\ps0.05 & 0.68\ps0.01 & 0.09\ps0.03 \\ \hline
    \multirow{6}{*}{\textbf{IPPM}} &\textit{Proportion} &\textit{0.76} &\textit{0.23} &\textit{0.09} &\textit{0.31} &\textit{0.24} &\textit{0.51} &\textit{0.67} &\textit{0.07} \\ \cline{2-10}
    &0-Shot & 0.85\ps0.02 & 0.05\ps0.05 & 0.09\ps0.06 & \textbf{0.51\ps0.02} & 0.42\ps0.02 & \textbf{0.59\ps0.02} & 0.76\ps0.03 & 0.12\ps0.05 \\ 
    &Theme & 0.85\ps0.02 & 0.45\ps0.05 & 0.20\ps0.06 & 0.49\ps0.01 & 0.41\ps0.00 & \textbf{0.59\ps0.02} & \textbf{0.77\ps0.00} & 0.15\ps0.04 \\ 
    &RAG & 0.77\ps0.05 & 0.05\ps0.02 & 0.08\ps0.10 & 0.47\ps0.00 & \textbf{0.46\ps0.05} & 0.58\ps0.03 & 0.75\ps0.02 & 0.11\ps0.03 \\ 
    &SFT & 0.86\ps0.00 & 0.08\ps0.02 & 0.08\ps0.08 & 0.32\ps0.03 & 0.38\ps0.05 & 0.49\ps0.03 & 0.73\ps0.01 & 0.07\ps0.07 \\ 
    &TADPOLE & \textbf{0.87\ps0.00} & \textbf{0.45\ps0.02} & \textbf{0.21\ps0.07} & 0.30\ps0.05 & 0.36\ps0.05 & 0.46\ps0.05 & 0.77\ps0.01 & \textbf{0.15\ps0.02} \\ \hline
    \multirow{8}{*}{\textbf{SyPPM}} &\textit{Proportion} &\textit{0.99} &\textit{0.79} &\textit{0.79} &\textit{0.33} &\textit{0.05} &\textit{0.75} &\textit{0.11} &\textit{0.56} \\ \cline{2-10}
    &0-Shot & 0.98\ps0.02 & 0.17\ps0.10 & 0.08\ps0.07 & \textbf{0.50\ps0.00} & 0.10\ps0.01 & 0.54\ps0.23 & 0.19\ps0.08 & 0.42\ps0.06 \\ 
    &Theme & \textbf{0.99\ps0.00} & 0.72\ps0.01 & 0.32\ps0.01 & 0.50\ps0.01 & 0.06\ps0.04 & 0.52\ps0.05 & 0.17\ps0.01 & 0.33\ps0.15 \\ 
    &RAG & 0.93\ps0.03 & 0.19\ps0.10 & 0.05\ps0.00 & 0.49\ps0.01 & 0.12\ps0.04 & 0.57\ps0.17 & 0.20\ps0.02 & 0.32\ps0.21 \\
    &SFT & 0.98\ps0.01 & 0.38\ps0.04 & 0.09\ps0.04 & 0.33\ps0.08 & \textbf{0.24\ps0.13} & \textbf{0.58\ps0.09} & \textbf{0.22\ps0.02} & 0.13\ps0.03 \\ 
    &TADPOLE & \textbf{0.99\ps0.00} & \textbf{0.79\ps0.03} & \textbf{0.49\ps0.01} & 0.16\ps0.04 & 0.17\ps0.09 & 0.19\ps0.02 & 0.22\ps0.03 & \textbf{0.46\ps0.16} \\
    &\textit{Gemini} &\textit{0.99} &\textit{0.71} &\textit{0.28} &\textit{0.50} &\textit{0.14} &\textit{0.76} &\textit{0.33} &\textit{0.71} \\ \hline
    &\textit{IAP} &\textit{0.80} &\textit{0.80} &\textit{0.53} &\textit{0.38} &\textit{0.07} &\textit{0.73} &\textit{0.15} &\textit{0.06} \\
    \hline
    \end{tabular}
    \caption{Class average theme-level edit-F1 scores for LLM adaptations. Each model adaptation is performed on three underlying LLMs, we report average results \ps standard deviation. We report micro average edit-F1 scores for each theme class. We also report the proportion of responses which contain each theme in each dataset. We include SyPPM results of the best commercial model (Gemini with theme prompting) for comparison. Finally, we report theme-level IAP, comparing LLM performance and expert human alignment at the theme level.}
    \label{app:tab:theme_classwise}
\end{table*}

In Tables \ref{tab:combined_results} and \ref{app:tab:combined_soc} we report content-level edit-F1 scores across the IPPM-SyPPM, and SoCPPM datasets, respectively. To investigate theme-specific performance of LLM response drafts, we also report theme class-specific scores at the content and theme levels. At the content level, in Table \ref{app:tab:content_classwise} we report the average recall of theme-labeled content within the expert responses of a given evaluation dataset. At the theme level, in Table \ref{app:tab:theme_classwise} we report the class-average edit-F1 scores when predicting expert response themes with LLM response draft themes. We discuss these results in Section \ref{sec:exp_results}.

In Table \ref{tab:combined_frontier} in Section \ref{sec:exp_results} we give content-level and theme-level edit-F1 scores for the Claude 4.5 Sonnet \cite{claude45sonnet}, Gemini 2.5 Pro \cite{comanici2025gemini25pushingfrontier} and GPT-OSS \cite{agarwal2025gpt} reasoning models. In Tables \ref{app:tab:frontier_content} and  \ref{app:tab:frontier_themewise} we similarly report the content-level average recall of theme-labeled content and the theme-level class-average edit-F1 scores.

\begin{table*}
  \centering
  \small
  \begin{tabular}{|c|c|cccccccc|}\hline
    \textbf{Prompt} & \textbf{Model} & \textbf{Emp} & \textbf{SymQ} & \textbf{MedQ} & \textbf{Assess} & \textbf{Plan} & \textbf{Logis} & \textbf{Coord} & \textbf{Cont} \\\hline
    \multirow{4}{*}{0-Shot} &GPT &\textbf{0.31} &0.07 &\textbf{0.07} &0.15 &0.21 &0.24 &0.24 &0.27 \\
    &Gemini &0.30 &\textbf{0.08} &\textbf{0.07} &0.15 &0.21 &\textbf{0.25} &0.24 &\textbf{0.28} \\
    &Claude  &\textbf{0.31} &0.07 &\textbf{0.07} &0.15 &\textbf{0.23} &0.24 &0.24 &0.27 \\
    &\textit{Avg} &\textit{0.31} &\textit{0.07} &\textit{0.07} &\textit{0.15} &\textit{0.22} &\textit{0.24} &\textit{0.24} &\textit{0.27} \\\hline
    \multirow{4}{*}{Theme} &GPT  &0.28 &0.07 &\textbf{0.08} &0.14 &\textbf{0.34} &0.21 &0.24 &0.19 \\
    &Gemini  &0.30 &0.07 &0.07 &\textbf{0.16} &0.23 &0.24 &0.24 &0.24 \\
    &Claude  &\textbf{0.31} &\textbf{0.08} &0.07 &0.14 &0.20 &0.24 &0.24 &\textbf{0.27} \\
    &\textit{Avg}  &\textit{0.30} &\textit{0.07} &\textit{0.07} &\textit{0.15} &\textit{0.26} &\textit{0.23} &\textit{0.24} &\textit{0.23} \\\hline
    \multicolumn{2}{|c|}{\textit{IAP}} &\textit{0.30} &\textit{0.21} &\textit{0.21} &\textit{0.13} &\textit{0.27} &\textit{0.37} &\textit{0.15} &\textit{0.64} \\
    \hline
    \end{tabular}
  \caption{Class average content-level recall scores for Claude 4.5 Sonnet, Gemini 2.5 Pro and GPT-OSS reasoning models models on the publicly-available SyPPM evaluation dataset. We evaluate each model using 0-shot and thematic prompts. Classifying elements in clinician responses into themes, we report response draft recall scores averaged across each theme. We also report content-level IAP, comparing LLM performance and expert human alignment at the content level.}
  \label{app:tab:frontier_content}
\end{table*}
\begin{table*}
  \centering
  \small
    \begin{tabular}{|c|c|cccccccc|}\hline
        \textbf{Prompt} & \textbf{Model} & \textbf{Emp} & \textbf{SymQ} & \textbf{MedQ} & \textbf{Assess} & \textbf{Plan} & \textbf{Logis} & \textbf{Coord} & \textbf{Cont} \\\hline
        \multirow{4}{*}{0-Shot} &GPT &\textbf{0.99} &0.42 &\textbf{0.22} &\textbf{0.50} &0.10 &\textbf{0.81} &0.27 &0.64 \\
        &Gemini &\textbf{0.99} &0.16 &0.03 &\textbf{0.50} &\textbf{0.12} &0.79 &0.29 &\textbf{0.66} \\
        &Claude &\textbf{0.99} &\textbf{0.49} &\textbf{0.22} &\textbf{0.50} &0.09 &0.53 &\textbf{0.34} &0.52 \\
        &\textit{Avg} &\textit{0.99} &\textit{0.36} &\textit{0.16} &\textit{0.50} &\textit{0.10} &\textit{0.71} &\textit{0.30} &\textit{0.61} \\\hline
        \multirow{4}{*}{Theme Prompting} &GPT &\textbf{0.99} &0.80 &\textbf{0.43} &\textbf{0.50} &0.10 &\textbf{0.82} &0.27 &0.60 \\
        &Gemini &\textbf{0.99} &0.71 &0.28 &\textbf{0.50} &\textbf{0.14} &0.76 &\textbf{0.33} &\textbf{0.71} \\
        &Claude &\textbf{0.99} &\textbf{0.87} &0.39 &\textbf{0.50} &0.12 &0.63 &0.15 &0.56 \\
        &\textit{Avg}  &\textit{0.99} &\textit{0.79} &\textit{0.37} &\textit{0.50} &\textit{0.12} &\textit{0.74} &\textit{0.25} &\textit{0.62} \\\hline
        \multicolumn{2}{|c|}{\textit{Theme Proportion}} &\textit{0.99} &\textit{0.79} &\textit{0.79} &\textit{0.33} &\textit{0.05} &\textit{0.75} &\textit{0.11} &\textit{0.56} \\\hline
        \multicolumn{2}{|c|}{\textit{IAP}} &\textit{0.80} &\textit{0.80} &\textit{0.53} &\textit{0.38} &\textit{0.07} &\textit{0.73} &\textit{0.15} &\textit{0.06} \\
        \hline
    \end{tabular}
  \caption{Class average theme-level edit-F1 scores for Claude 4.5 Sonnet, Gemini 2.5 Pro and GPT-OSS reasoning models models on the publicly-available SyPPM evaluation dataset. We evaluate each model using 0-shot and thematic prompts. We report edit-F1 scores for each theme class. Additionally, we report the proportion of responses which contain each theme (theme proportion) in the SyPPM dataset. Finally, we also report theme-level IAP, comparing LLM performance and expert human alignment at the theme level.}
  \label{app:tab:frontier_themewise}
\end{table*}


\section{Example REDCap Survey}
\label{app:redcap}

\begin{figure*}
    \centering
    \includegraphics[width=1.0\textwidth]{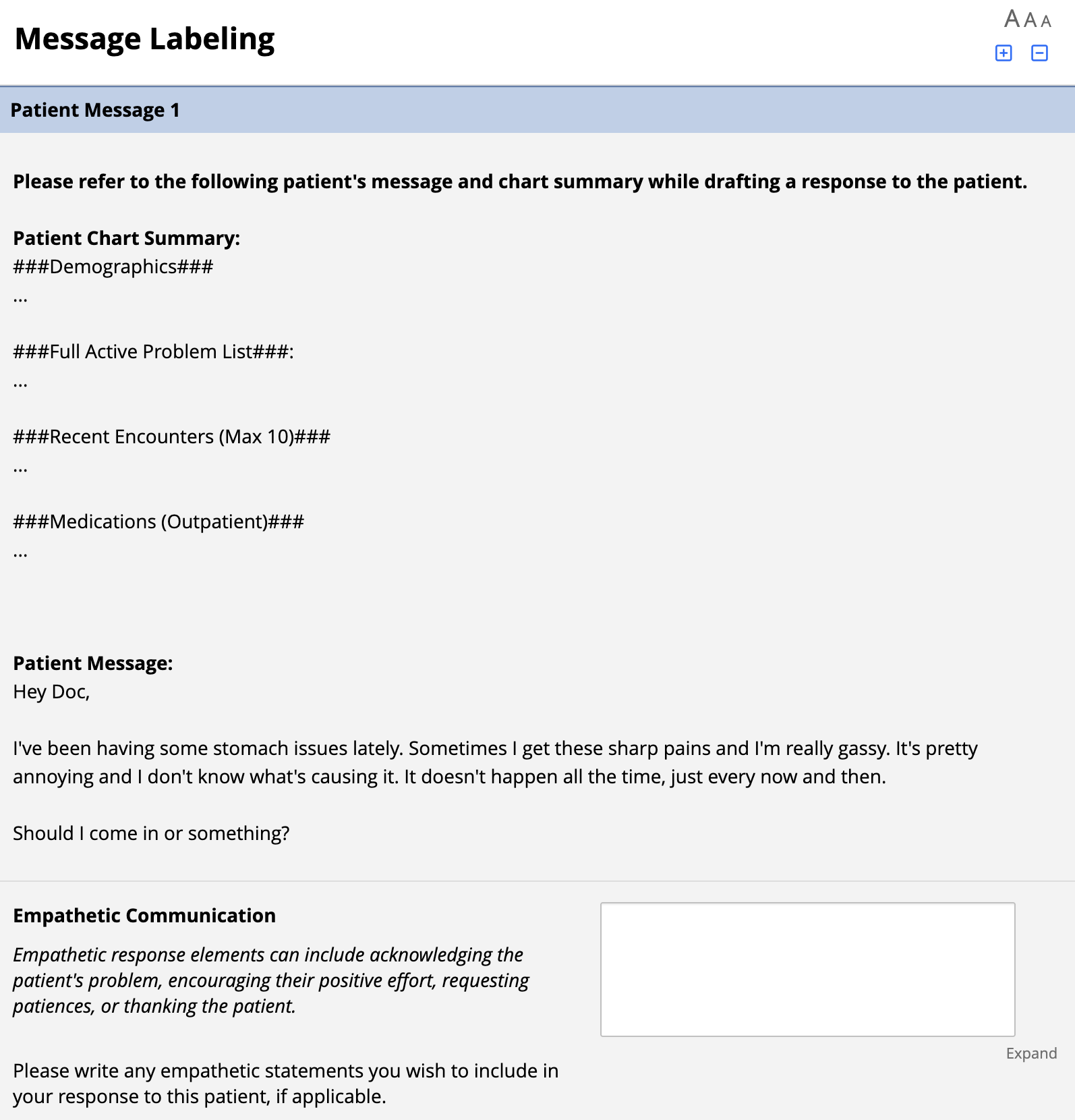}
    \caption{Screenshot of the beginning of a REDCap survey question used to collect clinician responses to patient messages in the SyPPM dataset. The patient's EHR chart and message are first given, then the clinician is prompted with a series of text entry boxes for each response theme described in Section \ref{sec:data_themes}.}
    \label{app:fig:redcap_1}
\end{figure*}

\begin{figure*}
    \centering
    \includegraphics[width=1.0\textwidth]{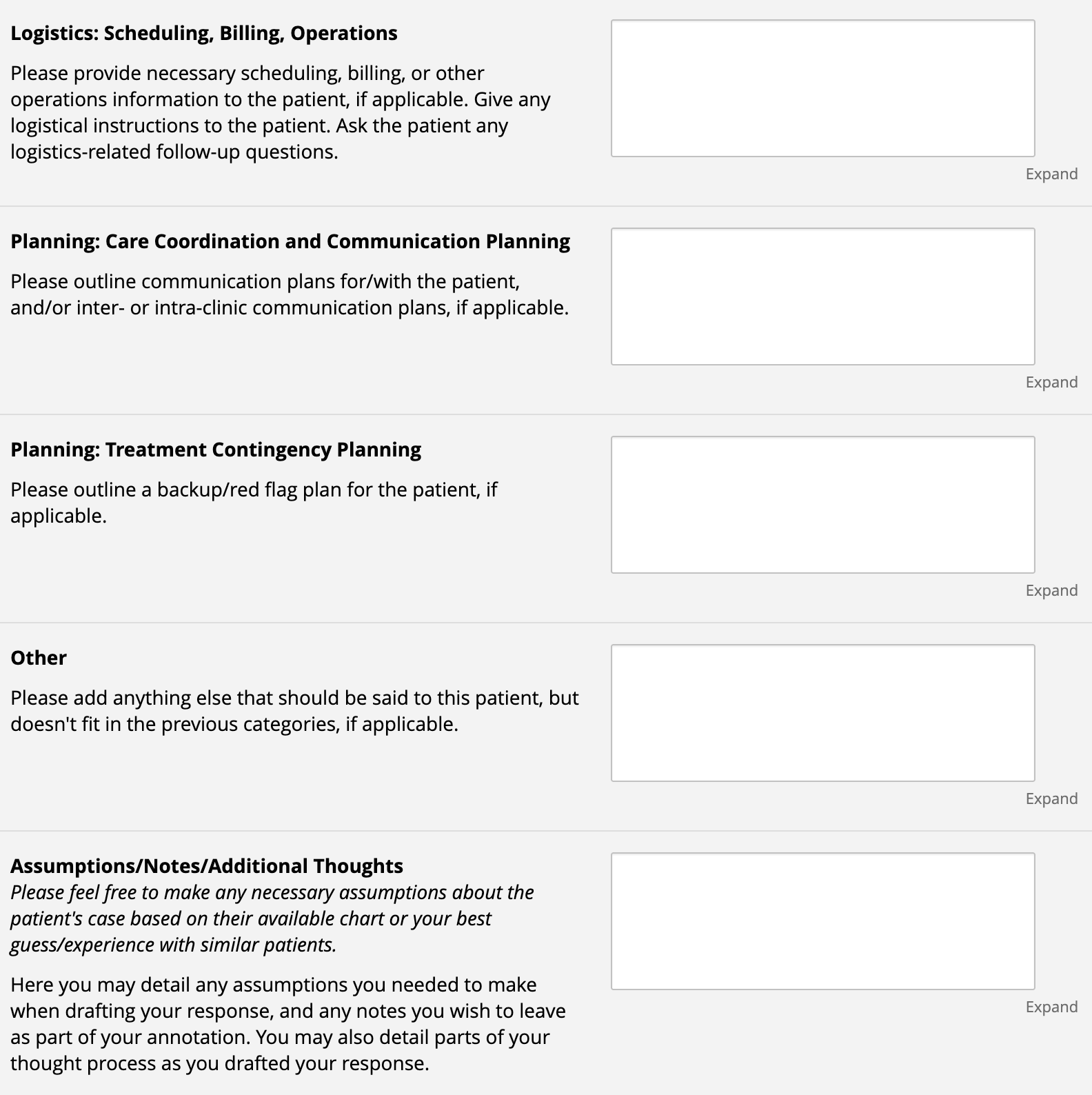}
    \caption{Screenshot of the end of a REDCap survey response used to collect clinician responses to patient messages in the SyPPM dataset. After seeing the patient's EHR chart and message, the clinician is prompted with a series of text entry boxes for each response theme described in Section \ref{sec:data_themes}. The clinician is also prompted to give any additional thoughts or assumptions they made while drafting their response.}
    \label{app:fig:redcap_2}
\end{figure*}

In the IPPM evaluation dataset, ground-truth responses are written by a paid team of 4 expert primary care nurses who work daily in the patient portal, collected via REDCap surveys \cite{harris2009research}. In the SyPPM evaluation dataset, ground-truth responses are written by a paid primary care doctor with 15+ years of experience. Each clinician was paid \$50 for every 10 responses (estimated to take 1 hour), in order to give ample time to write a full response to each message/EHR summary. While writing responses, experts were prompted ``if you had unlimited time, what would be included in your response to this patient?'' To provoke quality responses, clinicians were given a separate text entry box for each of the themes derived in Section \ref{sec:data_themes}. For example, the \textit{Treatment Contingency Planning} text box included the prompt ``please outline a backup/red flag plan for the patient, if applicable.'' Screenshots of an example REDCap survey question can be found in Figure \ref{app:fig:redcap_1} and Figure \ref{app:fig:redcap_2}.

\section{Prompts}
\label{app:prompts}

In Section \ref{sec:evaluation} we describe several methods for adapting LLMs for the patient message response drafting task. We give the 0-shot and thematic prompts in Figure \ref{app:fig:zshot_prompt} and Figure \ref{app:fig:theme_prompt}, respectively. The thematic prompt guides the model to use our derived themes when drafting responses to patient messages. In Section \ref{sec:exp_results} we see that thematic prompting, and other forms of contextual adaptation such as RAG, SFT, and our novel TADPOLE DPO-based strategy, improve LLM performance on the response drafting task.

\begin{figure*}[t] 
\centering
\begin{fullwidthbox}
\subfile{prompts/zshot_prompt}
\end{fullwidthbox}
\caption{0-shot prompt for patient message response drafting}
\label{app:fig:zshot_prompt}
\end{figure*}

\begin{figure*}[t] 
\centering
\begin{fullwidthbox}
\subfile{prompts/theme_prompt}
\end{fullwidthbox}
\caption{Thematic prompt for patient message response drafting}
\label{app:fig:theme_prompt}
\end{figure*}



\end{document}